\documentclass[10pt,twocolumn,letterpaper]{article}

\usepackage[pagenumbers]{cvpr} % To force page numbers, e.g. for an arXiv version
\makeatletter
\@namedef{ver@everyshi.sty}{}
\makeatother
\usepackage{graphicx}
\usepackage{amsmath}
\usepackage{amssymb}
\usepackage{booktabs}
\usepackage{multirow}
\usepackage{amsfonts}
\usepackage[table]{xcolor}
\usepackage{pgf}
\usepackage{subcaption}
\usepackage{listings}

\definecolor{yellow}{rgb}{1,1, 0.6}
\definecolor{orange}{rgb}{1, 0.8, 0.6}
\definecolor{red}{rgb}{1, 0.6, 0.6}

\usepackage[pagebackref,breaklinks,colorlinks]{hyperref}

\usepackage[capitalize]{cleveref}
\crefname{section}{Sec.}{Secs.}
\Crefname{section}{Section}{Sections}
\Crefname{table}{Table}{Tables}
\crefname{table}{Tab.}{Tabs.}

\def\cvprPaperID{1538} % *** Enter the CVPR Paper ID here
\def\confName{CVPR}
\def\confYear{2022}

\begin{document}

%%%%%%%%% TITLE - PLEASE UPDATE
\title{LTT-GAN: Looking Through Turbulence by Inverting GANs}
\author{Kangfu Mei \qquad Vishal M. Patel \\
Department of Electrical and Computer Engineering \\
Johns Hopkins University\\
{\tt\small \url{https://kfmei.page/LTT-GAN/}}
}

\twocolumn[{%
\renewcommand\twocolumn[1][]{#1}%
\vspace{-4\baselineskip}
\maketitle
\vspace{-3\baselineskip}
\begin{center}
    \centering
    \setlength{\tabcolsep}{0.5pt}
    \captionsetup{type=figure}
    {\small
    \renewcommand{\arraystretch}{0.5} 
    \begin{tabular}{c c c c c c c c c}
    \raisebox{0.3in}{\rotatebox[origin=t]{90}{Input}}&
    \includegraphics[width=0.12\linewidth]{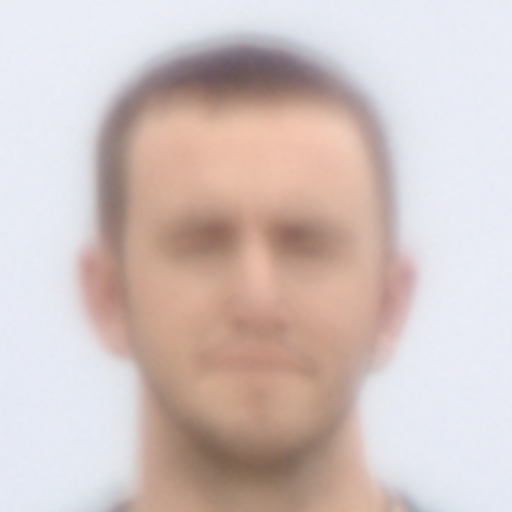}&
    \includegraphics[width=0.12\linewidth]{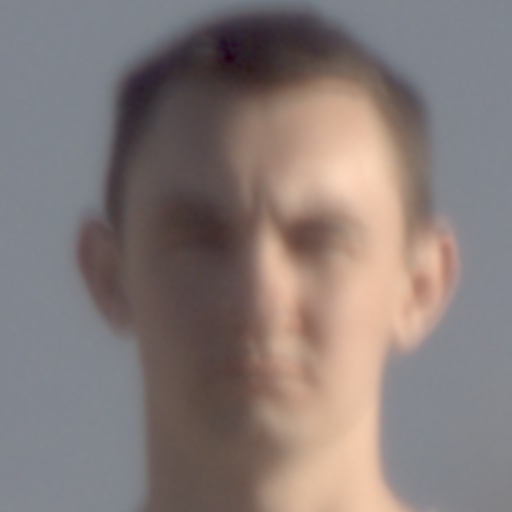}&
    \includegraphics[width=0.12\linewidth]{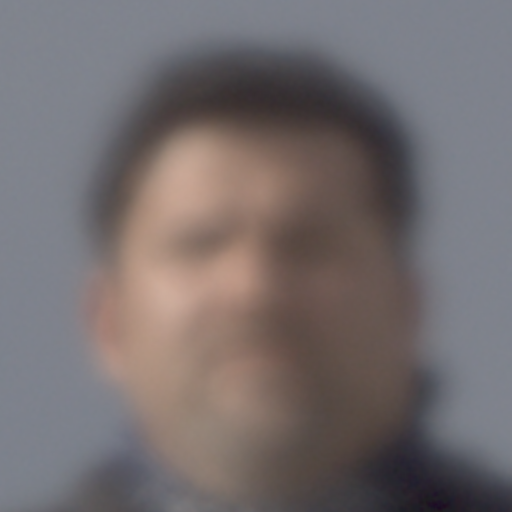}&
    \includegraphics[width=0.12\linewidth]{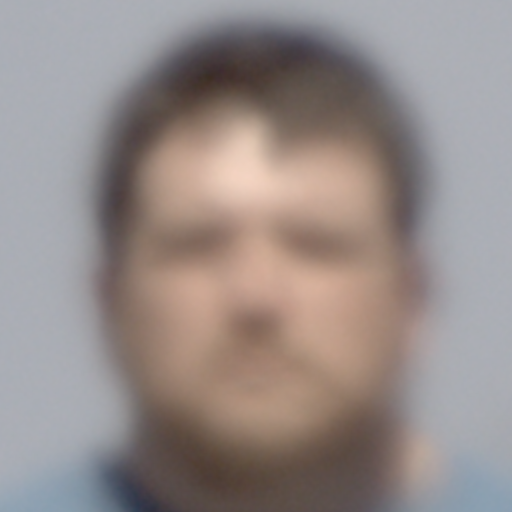}&
    \includegraphics[width=0.12\linewidth]{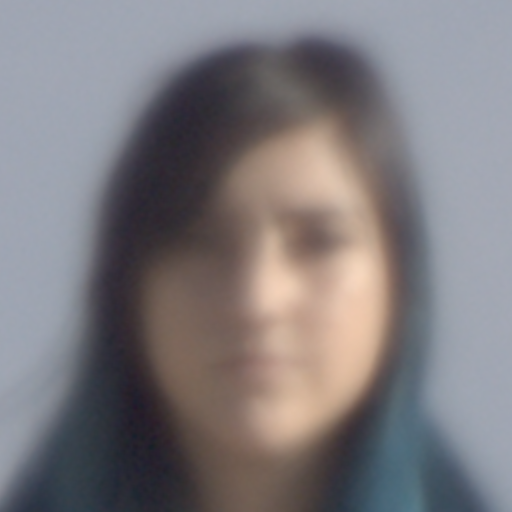}&
    \includegraphics[width=0.12\linewidth]{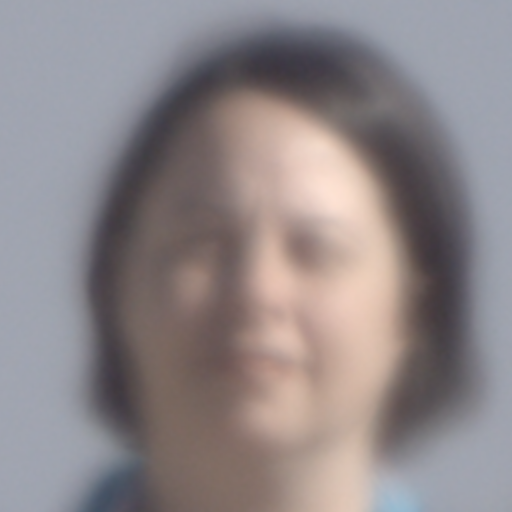}&
    \includegraphics[width=0.12\linewidth]{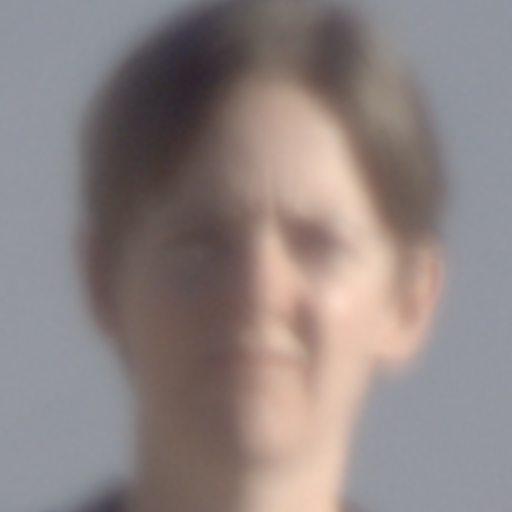}&
    \includegraphics[width=0.12\linewidth]{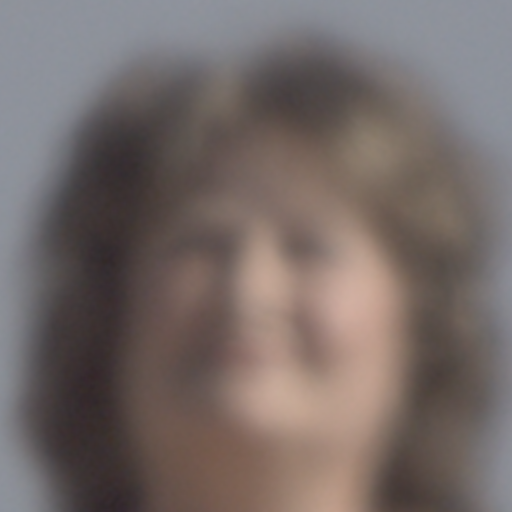}
    \tabularnewline
    \raisebox{0.32in}{\rotatebox[origin=t]{90}{SOTA~\cite{lau_atfacegan_2021}}}&
    \includegraphics[width=0.12\linewidth]{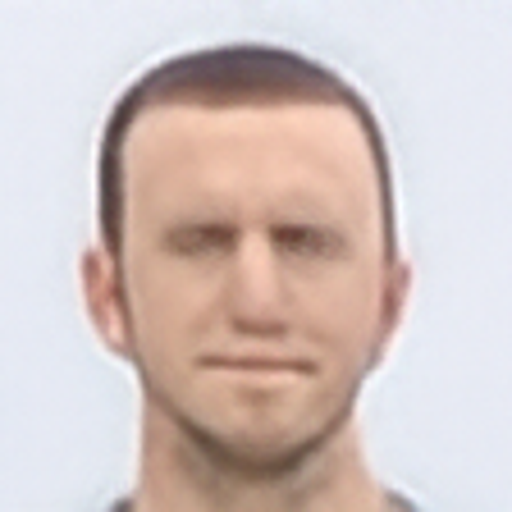}&
    \includegraphics[width=0.12\linewidth]{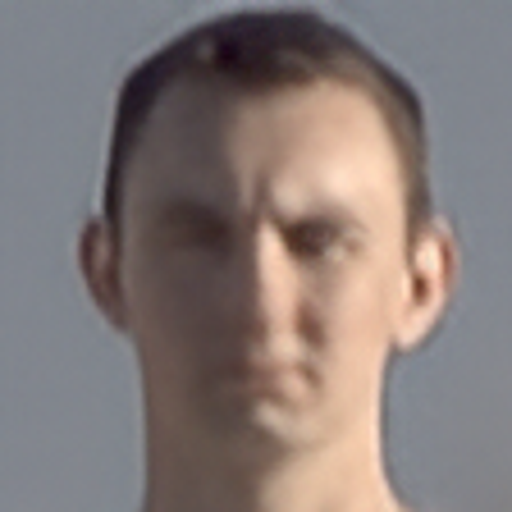}&
    \includegraphics[width=0.12\linewidth]{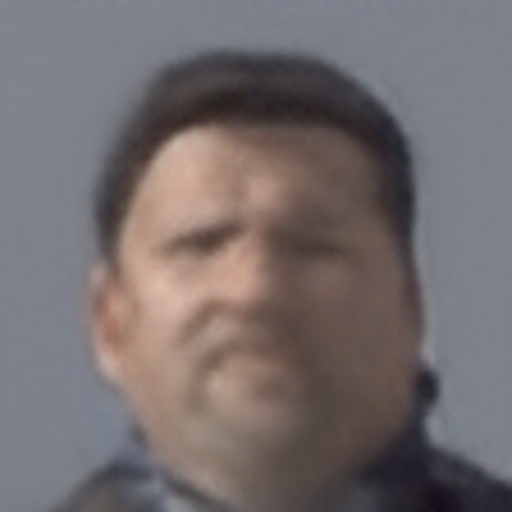}&
    \includegraphics[width=0.12\linewidth]{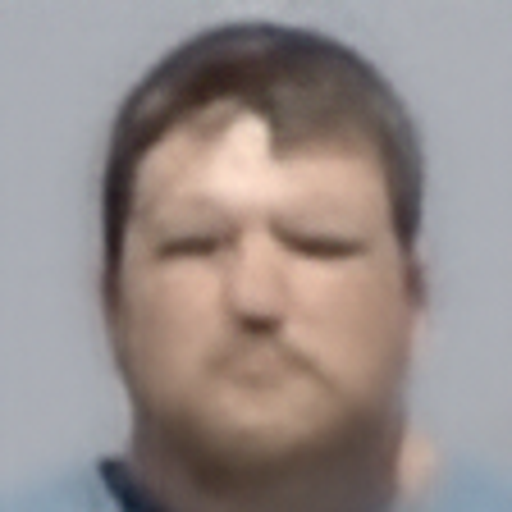}&
    \includegraphics[width=0.12\linewidth]{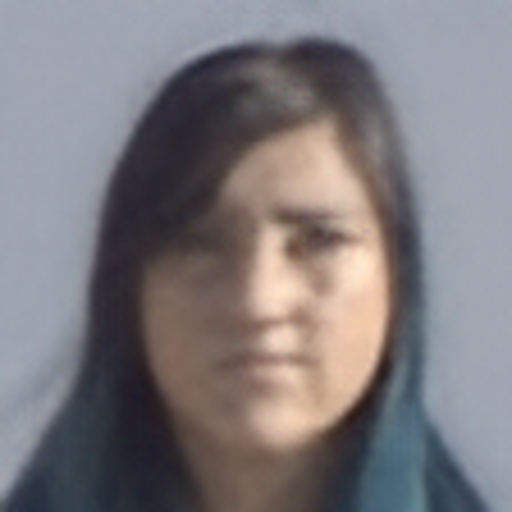}&
    \includegraphics[width=0.12\linewidth]{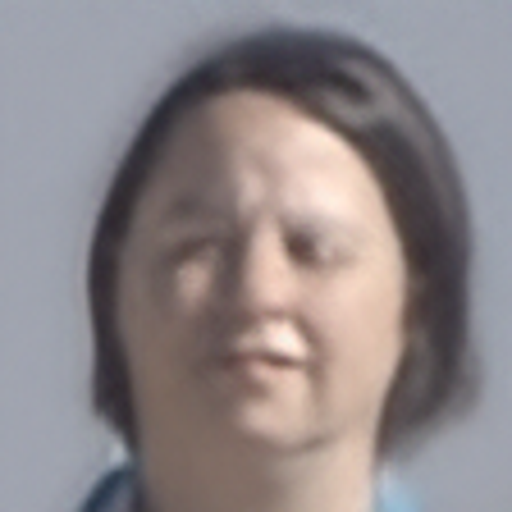}&
    \includegraphics[width=0.12\linewidth]{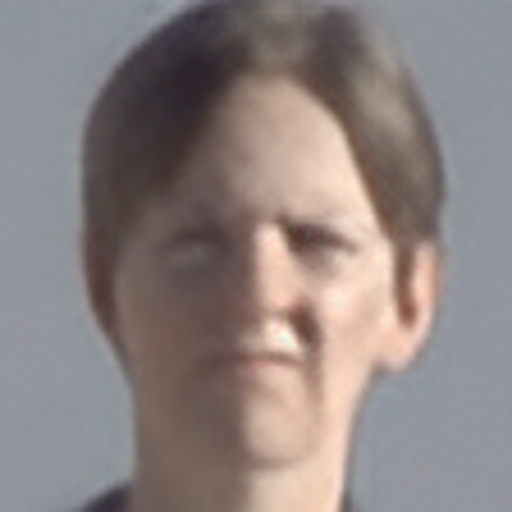}&
    \includegraphics[width=0.12\linewidth]{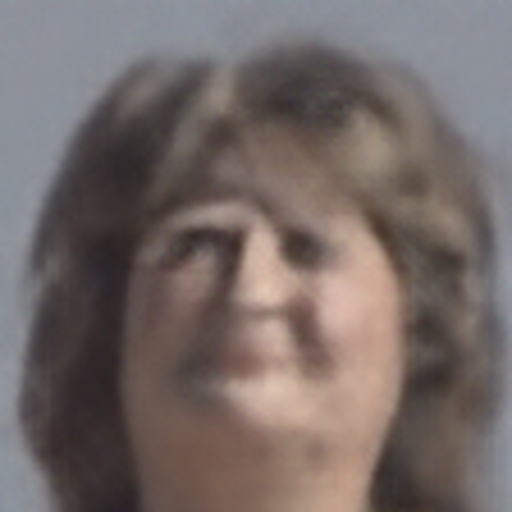}
    \tabularnewline
    \raisebox{0.3in}{\rotatebox[origin=t]{90}{Ours}}&
    \includegraphics[width=0.12\linewidth]{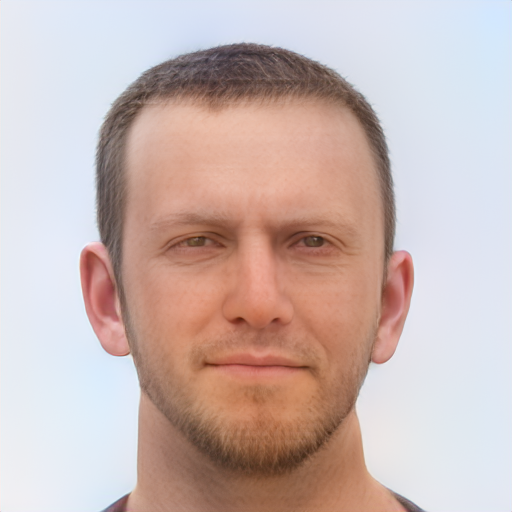}&
    \includegraphics[width=0.12\linewidth]{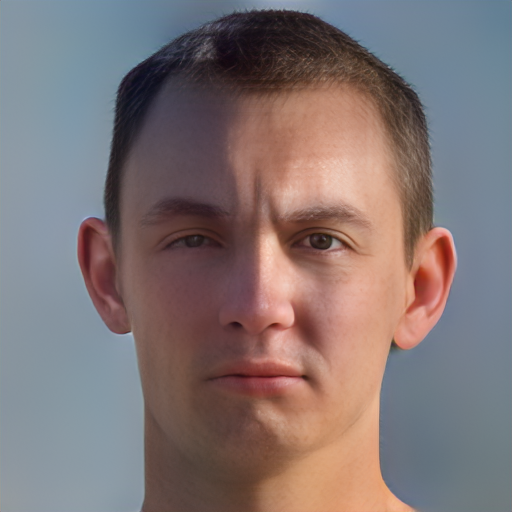}&
    \includegraphics[width=0.12\linewidth]{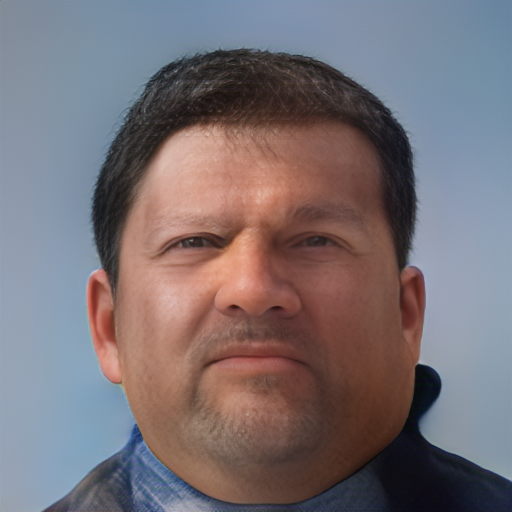}&
    \includegraphics[width=0.12\linewidth]{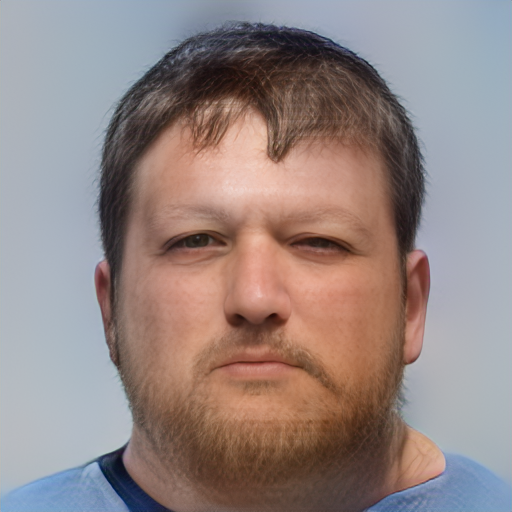}&
    \includegraphics[width=0.12\linewidth]{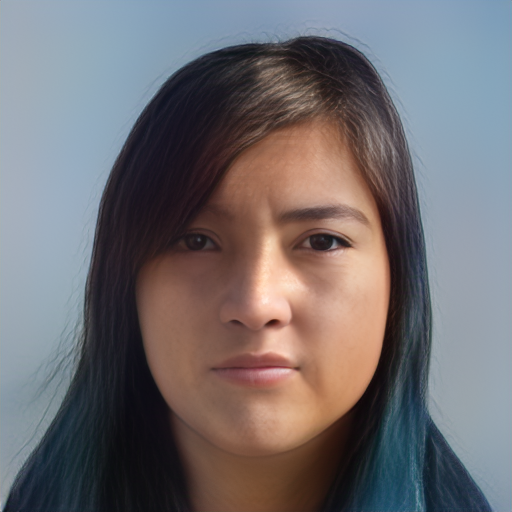}&
    \includegraphics[width=0.12\linewidth]{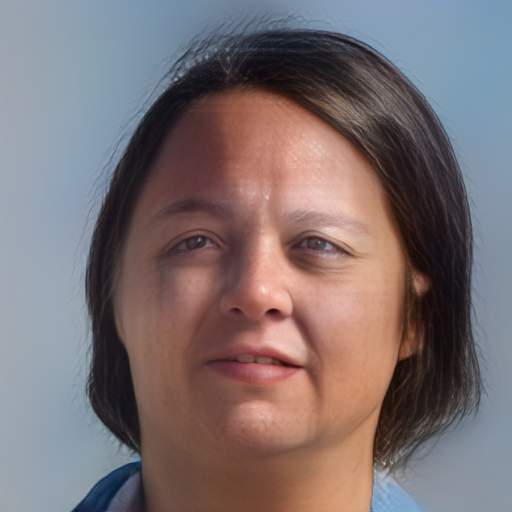}&
    \includegraphics[width=0.12\linewidth]{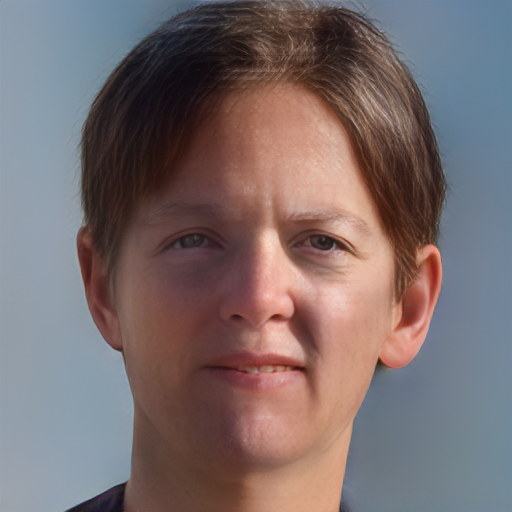}&
    \includegraphics[width=0.12\linewidth]{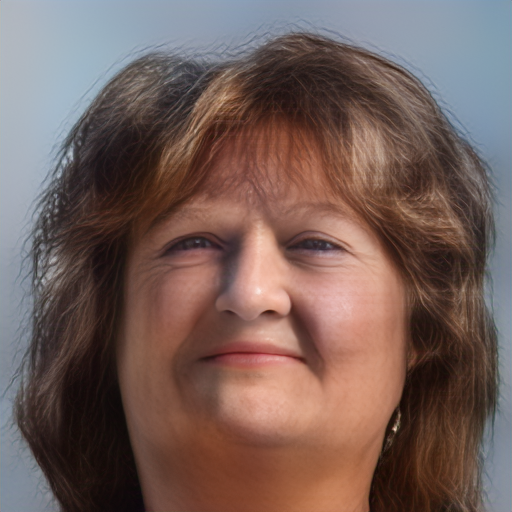}
    \tabularnewline
    \raisebox{0.3in}{\rotatebox[origin=t]{90}{Reference}}&
    \includegraphics[width=0.12\linewidth]{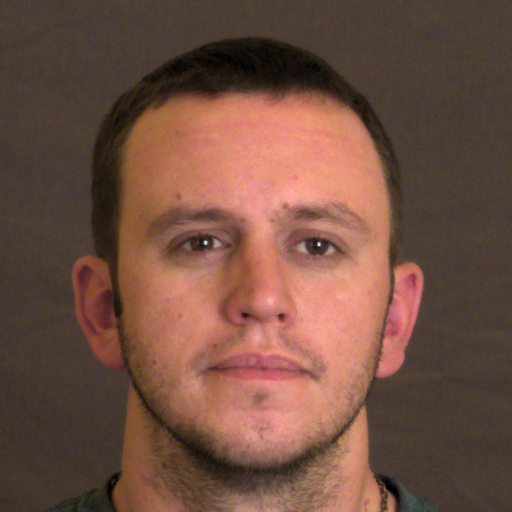}&
    \includegraphics[width=0.12\linewidth]{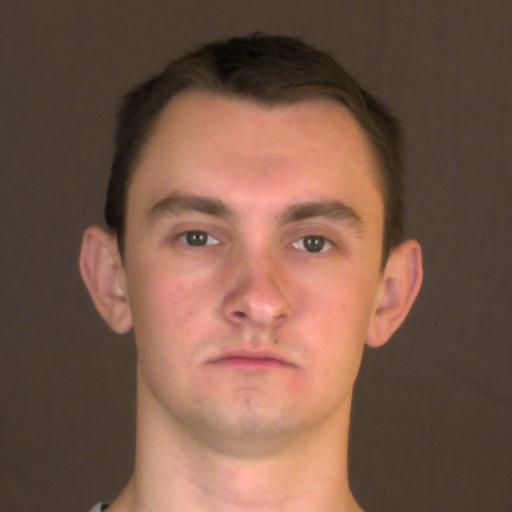}&
    \includegraphics[width=0.12\linewidth]{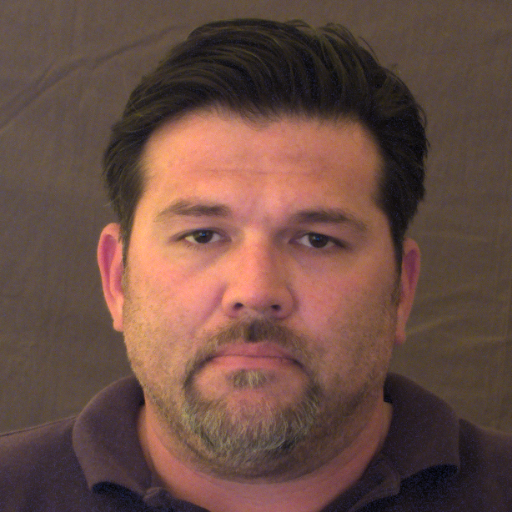}&
    \includegraphics[width=0.12\linewidth]{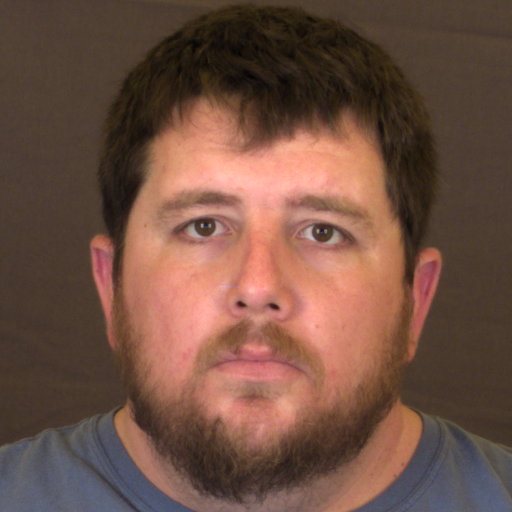}&
    \includegraphics[width=0.12\linewidth]{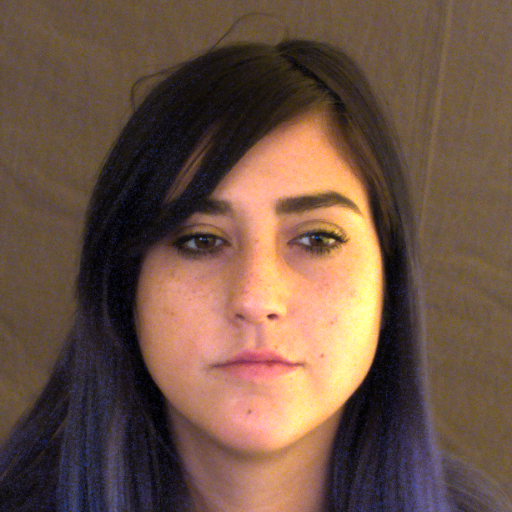}&
    \includegraphics[width=0.12\linewidth]{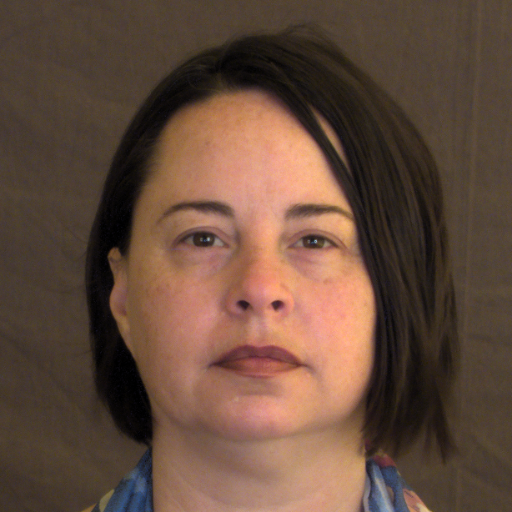}&
    \includegraphics[width=0.12\linewidth]{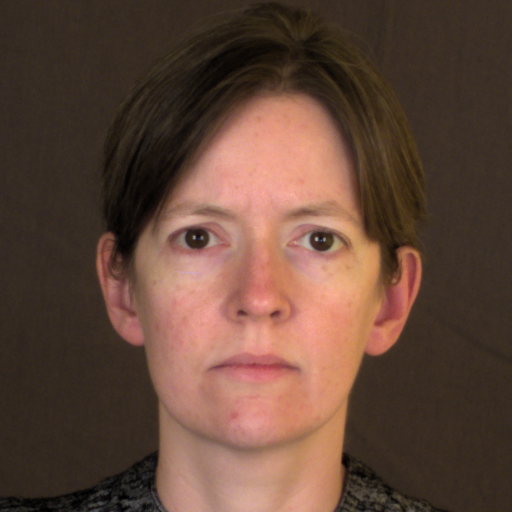}&
    \includegraphics[width=0.12\linewidth]{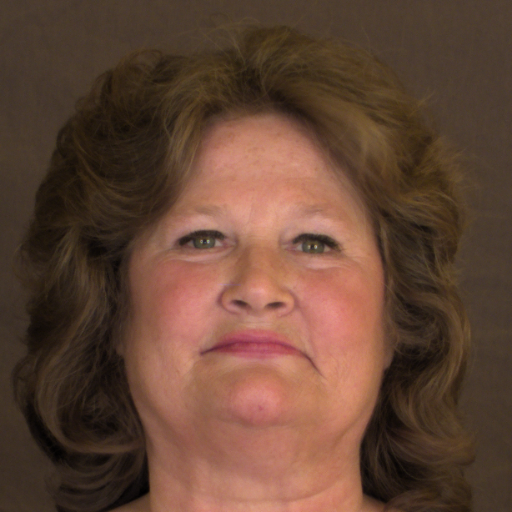}
    \end{tabular}}
    \vspace{-0.5\baselineskip}
    \captionof{figure}{First row: Atmospheric turbulence degraded images in 300 meters range. Second row: Results corresponding to previous state-of-the-art (SOTA) method~\cite{lau_atfacegan_2021}.  Third row: Results corresponding to the proposed method.  Last row: reference images without turbulence.
    The proposed method is able to produce sharper results compared to the previous SOTA method.
    All images are of resolution $512\times 512$.}
\end{center}%
}]

% \maketitle

%%%%%%%%% ABSTRACT
\begin{abstract}
In many applications of long-range imaging, we are faced with a scenario where a person appearing in the captured imagery is often degraded by atmospheric turbulence.
However, restoring such degraded images for face verification is difficult since the degradation causes images to be geometrically distorted and blurry.
To mitigate the turbulence effect, in this paper, we propose the first turbulence mitigation method that makes use of visual priors encapsulated by a well-trained GAN.
Based on the visual priors, we propose to learn to preserve the identity of restored images on a spatial periodic contextual distance.
Such a distance can keep the realism of restored images from the GAN while considering the identity difference at the network learning.
In addition, hierarchical pseudo connections are proposed for facilitating the identity-preserving learning by introducing more appearance variance without identity changing.
Extensive experiments show that our method significantly outperforms prior art in both the visual quality and face verification accuracy of restored results.
\end{abstract}

%%%%%%%%% BODY TEXT
\section{Introduction}
\label{sec:intro}
Generative Adversarial Network (GAN) inversion is a new trending way in image restoration, which has been applied in applications, including but not limited to \emph{image super-resolution}~\cite{menon_pulse_2020, chan_glean_2021}, \emph{image colorization}~\cite{wu_towards_2021}, \emph{blind face restoration}~\cite{yang_gan_2021,wang_towards_2021}, and \emph{general tasks}~\cite{gu_image_2020, pan_exploiting_2021}.
Despite photo-realistic results can be restored from well-trained GANs, inverting degraded images is nontrivial.
Among major inversion ways~\cite{zhu_generative_2016,menon_pulse_2020,gu_image_2020, pan_exploiting_2021, lin2021anycost, pidhorskyi2020adversarial, richardson_encoding_2021, yang_gan_2021, wang_towards_2021, chan_glean_2021, tov_designing_2021}, the learning based method achieves the most compelling efficiency and does not need the awareness of the degradation.
Although it is practical, unfaithful realism and unnatural details can be observed in its results, especially in the cases where large degradations exist \eg, atmospheric turbulence.

Learning-based methods, which we call \emph{Generative Embedding Network (GEN)}, usually consist of a learnable latent code predictor and an embedded well-trained GAN~\cite{brock2018large, karras_style-based_2019}.
They can transform the predicted latent code from the degraded image into a clear image with realistic details with the help of a well-trained GAN.
Among them, the latent code predictor of GEN learns with an additional pixel-wise loss between the restored image and the target image, \eg, $\mathcal{L}_2$, together with the original adversarial loss, for preserving the identity and realism of restored results, respectively.
Under such a scenario, the motivation of our method comes from three different aspects theoretically and empirically.

a) \emph{Theoretically, the term of applied adversarial loss and pixel-wise loss is statistically different.}
We assume that this difference leads to unfaithful realism of restored results.
Thus, we take inspiration from the contextual approach~\cite{mechrez_contextual_2018} and replace the pixel-wise loss with it, for preserving identity while reducing the statistical difference.
By comparing regions with similar semantics only, the contextual loss can bypass the alignment regularization of standard pixel-wise losses.
Moreover, such a loss maintains statistical difference of images without explicitly estimating the density of high-dimensional semantic features, which can be approximated to the Kullback–Leibler (KL) divergence between two images~\cite{mechrez_maintaining_2018}.
Such a term is actually similar to the adversarial loss that can be approximated to the Jensen–Shannon (JS) divergence between two images~\cite{arjovsky_towards_2017}.
We empirically find that simply replacing the $\mathcal{L}_2$ loss with the contextual loss in GEN learning achieves 1.77\% improvement on the facial embedding cosine distance.

b) \emph{Empirically, the fine details (\eg, eyes) related to the identity are difficult to restore exactly only if the coarse appearance (\eg, color and pose) is exactly restored.}
We argued that this results from the hierarchical generation process used by the embedded GAN, \eg, StyleGAN~\cite{karras_style-based_2019}.
It combines interpolated multi-scale coarse features to generate fine details of restored images.
However, error always exists in the coarse features, and thus the entangled fine detail cannot be easily learned to be preserved.
We further argue that the identical features of each image is redundant, and thus delicately extracting sub-images from the original one does not change its identity.
By comparing the statistical context of sub-images with the same identity but different appearances, our new contextual distance can tolerate the differences of coarse appearance and better consider the identity differences between the restored results and the ground truth.
Such a modification on the original contextual distance leads to 2.40\% improvement.

c) \emph{The network should produce multiple results from a single input. Because multiple clean images can produce the same degraded image.}
We further take inspiration from \emph{b)} and empirically find that gradually changing coarse features (\eg, features in the shallow layer of StyleGAN) can result images with the same identity but different coarse appearance, which can be applied to implicitly correspond to multiple ground truth.
Specifically, we connect the hierarchical layers of the embedded StyleGAN with multiple modulation features.
By repeating hierarchical layers with different modulation features, the embedded StyleGAN can produce multiple identical images in a single forward pass.
Comparing sub-images with more appearance variations on the modified contextual distance with the modified network as the loss further achieves 3.77\% improvement.

Overall, our method is based on the recent state-of-the-art GEN method GFPGAN~\cite{wang_towards_2021}, but instead learns by a new contextual manner illustrated in Figure~\ref{fig:pcp}, and with a different hierarchical generation manner illustrated in Figure~\ref{fig:pipe}.
The effectiveness of ours is shown in the turbulence face image mitigation problem, illustrated in Figure~\ref{fig:turbulence}.
Note that this is the first method that can produce sharp-looking face images from turbulence degraded imagery in a resolution of $512 \times 512$.
Our method substantially improves upon the previous best method by 4.51 in visual quality metric \emph{FID} and recognition accuracy \emph{Top1} of 11.24\%, on synthetic and real-world turbulence degraded face images, respectively.

\section{Related Works}
\paragraph{Atmospheric Turbulence Simulation and Mitigation}  based on deep neural networks has  gained traction in recent years due to its applications in long-range surveillance and biometrics.  A detailed discussion regarding turbulence simulation and its effects on images can be found in ~\cite{roggemann_imaging_2018, goodman_statistical_2015}.
For training a deep network for restoring images degraded by turbulence, Chak et al.~\cite{chak_subsampled_2018} and Lau et al.~\cite{lau_subsampled_2021} proposed random distortions and blur-based on handcrafted rules to synthesize pair-wise turbulence degraded images.  This data augmentation/synthesis method has been successfully used in many restoration works \cite{lau_atfacegan_2021}, \cite{yasarla_learning_2021}, \cite{nair_confidence_2021}.
More recent works from Chimitt et al.~\cite{chimitt_simulating_2020} and its improved version by Mao et al.~\cite{mao2021accelerating} follow the split-step propagation method to simulate the effect of turbulence on images~\cite{bos_technique_2012, hardie_simulation_2017}.
It can model the turbulence caused by wavefront distortion, and it statistically fits the theoretical predictions of turbulence.
In this paper, we use the simulation method from Mao et al.~\cite{mao2021accelerating} to synthesize turbulence degraded images.

\vspace{-1\baselineskip}
\paragraph{GAN Inversion for Restoration} with well-trained GANs can provide photo-realistic priors, and thus it is a new promising avenue for image restoration.
According to the recent survey of GAN inversion~\cite{xia_gan_2021}, there are two major ways for embedding a well-trained GAN into the restoration framework.
Typical methods~\cite{creswell_inverting_2018, ma_invertibility_2018, gu_image_2020, menon_pulse_2020} belong to the first approach in which the optimal latent code of GANs are retrieved with some iterative optimization algorithms. Different modifications on the latent space definition or optimization procedure are proposed.
Methods such as ~\cite{zhu_generative_2016,bau_seeing_2019, yang_gan_2021, chan_glean_2021} fall in the second approach, in which an additional encoder is employed to predict the optimal point, which usually requires pair-wise data to train the additional encoder with pixel-wise loss.
Our method falls in the second category, but our learning objective differs from these typical methods and can better consider the identity of restored results for restoration learning.

\section{Proposed Method}
\subsection{Preliminaries}
\begin{figure}[t]
  \centering
  \includegraphics[width=\linewidth]{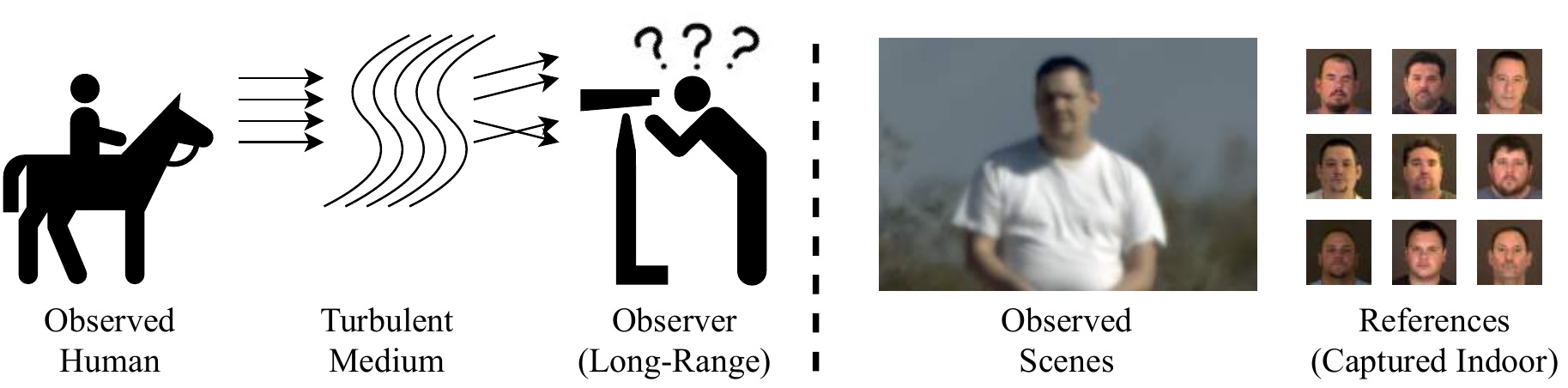}
  \vspace{-1.5\baselineskip}
  \caption{Atmospheric turbulence causes geometric distortion and blur effects on the captured images that disturbs face recognition.}
  \vspace{-1\baselineskip}
  \label{fig:turbulence}
\end{figure}

Following ~\cite{yasarla_learning_2020, lau_atfacegan_2021, yasarla_learning_2021, nair_confidence_2021}, the degradation due to atmospheric turbulence can be expressed as follows
\setlength{\belowdisplayskip}{2pt} \setlength{\belowdisplayshortskip}{2pt}
\setlength{\abovedisplayskip}{2pt} \setlength{\abovedisplayshortskip}{2pt}
\begin{equation}\label{eq:turb}
  \tilde{I}_k = D_k(H_k(I)) + n_k,
\end{equation}
where $\tilde{I}_k$ is the single turbulence image, $D_k(*)$ and $H_k(*)$ are the deformation operator and turbulence-induced point spread function, respectively, $I$ is the clean image, and $n_k$ is the noise.
Both degradations in Equation~\eqref{eq:turb} can be combined into a single degradation $T(\cdot)$ and as a result, we obtain the following more general observation model
\begin{equation}
  \tilde{I} = T(I).
\end{equation}
Turbulence mitigation entails finding  an inverse function that can map arbitrary turbulence degraded image $\tilde{I}$ into image $G(\tilde{I})$ that is close to the clear image $I$, as shown in Figure~\ref{fig:turbulence}.  However, this is an  ill-posed problem due to the large variations in degradation $T(\cdot)$.  In other words, each turbulence degraded image $\tilde{I}$ can correspond to multiple clear images.
Here, we define these possible ``ground truths" as an image set $\textbf{I} = \{I_1,I_2, \dots, I_n\}$ corresponding to the restored image $G(\tilde{I})$.
Conventional pixel-wise loss, \eg, $\mathcal{L}_{2}$ used in most GEN networks can be written as
\begin{equation}
  \label{eq:mse}
  \mathcal{L}_{MSE} (G) = \frac{1}{n} \sum^n_i \| G(\tilde{I}) - I_i \|^2, ~I_i \in \mathbf{I}.
\end{equation}
According to the derivation~\cite{yang2019deep}, minimizing Equation~\eqref{eq:mse} is equivalent to minimize the maximum likelihood estimation of the conditional empirical distribution $\mathbb{C}_g$ of $G(\tilde{I}) | \tilde{I}$ and the averaged conditional distribution $\mathbb{C}_{\mathbf{I}}$ of $\mathbf{I} | \tilde{I}$ as
\begin{equation}
  \label{eq:avg}
  \mathcal{L}(G) = \mathbb{E}_{x \sim \mathbb{C}_{\mathbf{I}}}  \left[\log\frac{\prod^n_i \{\mathbb{C}_{I_i|\tilde{I}}(x)\}}{\mathbb{C}_g(x)}\right].
\end{equation}
The averaged loss term will perform differently from the original one depending on the variation level of degradations.
Thus, the difference is usually ignored in tasks with limited variance, \eg, super-resolution.

In contrast, GANs are able to generate realistic looking images due to the use of adversarial loss, \eg, the logistic loss function with a discriminator $D(\cdot)$.
When learning a GEN, which fine-tunes the network with a well-trained GAN, the applied well-trained discriminator is optimal.
Thus, the adversarial learning of the network $G(\cdot)$ is equivalent to minimize the Jensen-Shannon divergence between the two image distributions according to~\cite{arjovsky_towards_2017} as
\begin{equation}
  \label{eq:gan}
  \begin{split}
    \mathcal{L}(D, G) &= 2\mathrm{JS}(\mathbb{P}_{\mathbf{I}} \| \mathbb{P}_{g}) -2\log2,\\
    \mathrm{JS}(\mathbb{P}_{\mathbf{I}} \| \mathbb{P}_{g}) & = \mathbb{E}_{x \sim \mathbb{P}_{\mathbf{I}}}  \left[\log\frac{\mathbb{P}_{\mathbf{I}}(x)}{\mathbb{P}_g(x)}\right] \\
    &+ \mathbb{E}_{x \sim \mathbb{P}_{\mathbf{g}}}  \left[\log\frac{\mathbb{P}_{g}(x)}{\mathbb{P}_{\mathbf{I}}(x)}\right],
  \end{split}
\end{equation}
where $\mathbb{P}_{\mathbf{I}}$ and $\mathbb{P}_g$ are the probability distributions of the ground truth and restored results, respectively.
According to our \emph{motivation a)}, the difference between Equation~\eqref{eq:mse} and Equation~\eqref{eq:gan} shows inconsistency in the gradient orientation of the network learning.
Therefore, it is difficult to achieve a good trade-off between realism and identity preservation in the conventional way of learning GENs, especially in the restoration tasks with large variations in the degradation (i.e. turbulence in this paper).

\vspace{-1\baselineskip}
\paragraph{Contextual Distance} 
From the derivation in~\cite{mechrez_maintaining_2018} corresponidng to the contextual loss~\cite{mechrez_contextual_2018}, which was originally proposed for misaligned data, the contextual distance in perceptual features $X$ and $Y$ of images can be approximated as
\begin{equation}
K L\left(\mathbb{P}_{X} \| \mathbb{P}_{Y}\right)=\int \mathbb{P}_{X} \log \frac{\mathbb{P}_{X}}{\mathbb{P}_{Y}},
\end{equation}
where $\mathbb{P}_{X} $ and $\mathbb{P}_{Y}$ are the density of points estimated on the perceptual features, with multivariate kernel density estimation.
Due to the similarity in terms of the adversarial and contextual losses, we empirically find that replacing $\mathcal{L}_2$ loss with the contextual loss in training a GEN can significantly accelerate the convergence.
However, there has been no demonstration that such a contextual loss is capable of preserving identity as well as realism of restored results.

\begin{figure*}[htbp]
  \center
  \includegraphics[width=\linewidth]{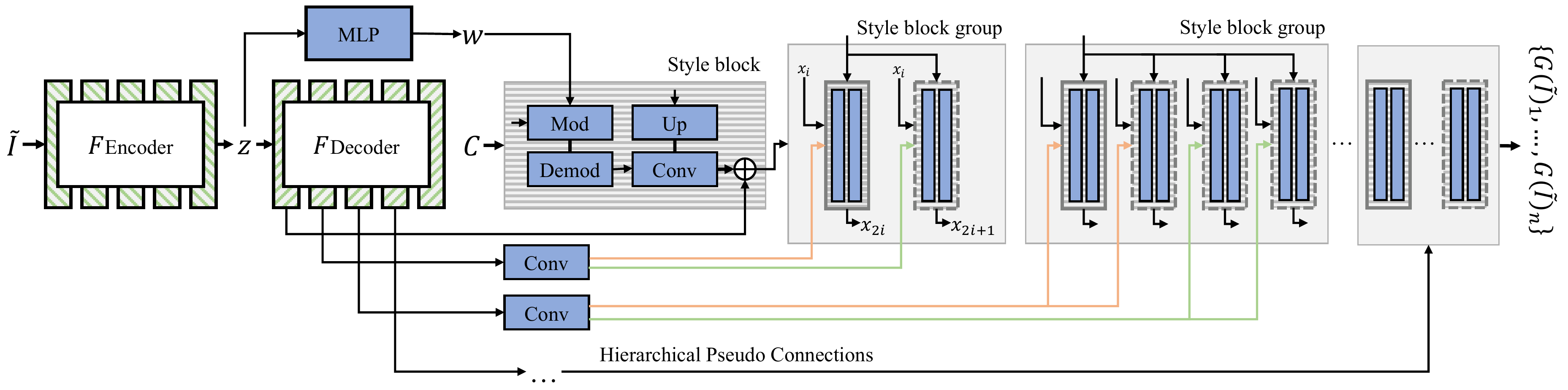}
  \vspace{-1.5\baselineskip}
  \caption{Overview of our method. Here, $F_{\mathrm{Encoder}}$ and $F_{\mathrm{Decoder}}$ take the turbulence degraded image $\tilde{I}$ as input, and generate the latent code $z$ and modulation features for the embedded StyleGAN. The performance of StyleGAN is improved by Hierarchical Pseudo Connections with multiple pseudo style blocks. Thus, the network produces $n$ final results instead of a single result in a single forward pass.}
  \label{fig:pipe}
  \vspace{-1\baselineskip}
\end{figure*}

\subsection{Spatial Periodic Contextual Distance}
\label{sec:pcp}
\begin{figure}[htbp]
  \center
  \includegraphics[width=1\linewidth]{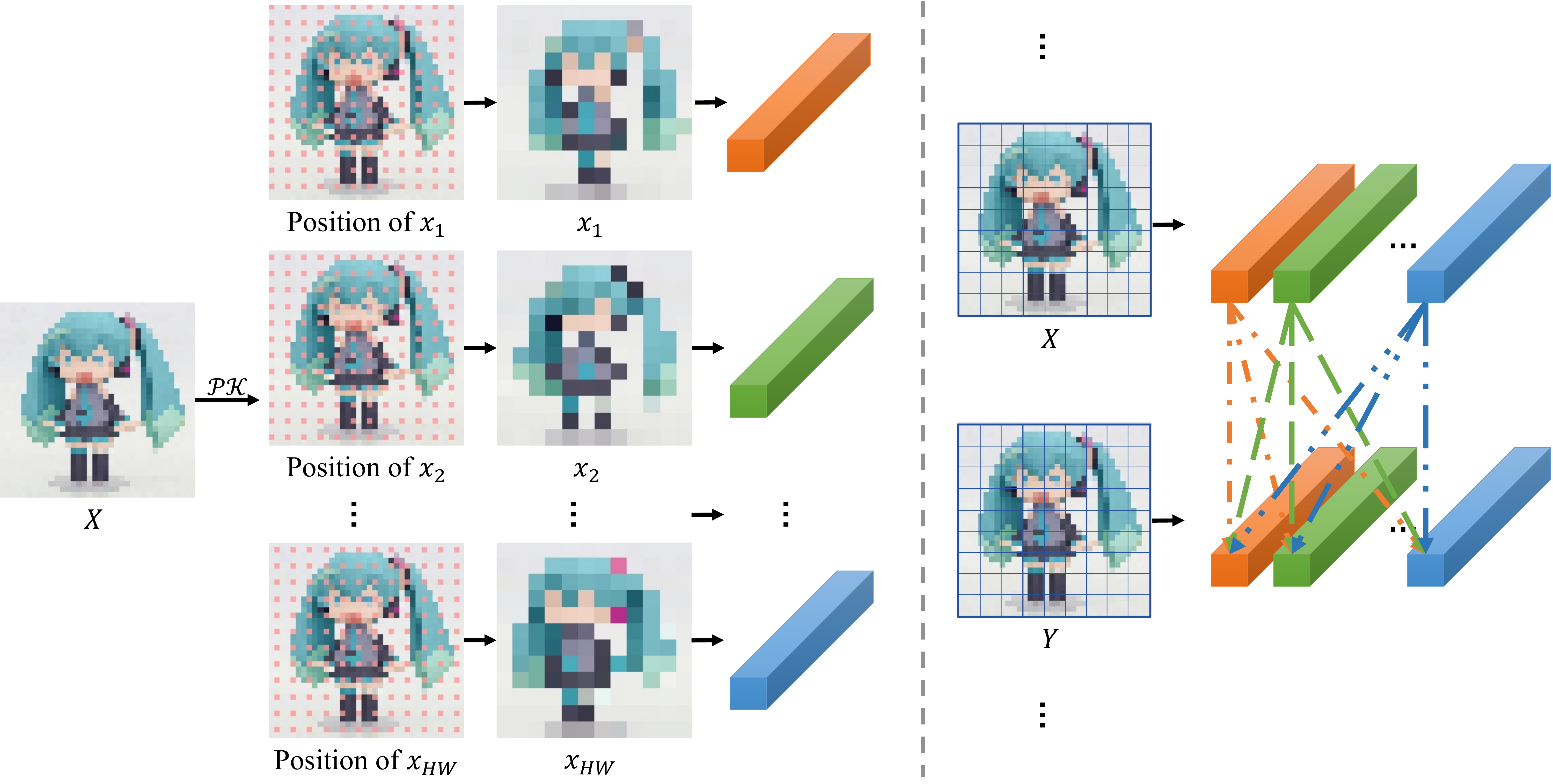}
  \vspace{-1.5\baselineskip}
  \caption{The procedure of $\mathcal{P}\mathcal{K}$ in a given image $X$. The sampled positions of each sub-image $x_i$ are highlighted in image $X$. These sub-images are then flattened into samples for contextual distance.}
  \label{fig:pcp}
  \vspace{-1.5\baselineskip}
\end{figure}

In our method we treat each image as a collection of multiple sub-images extracted spatial periodically, and we then consider the contextual distance between the two sub-image collections as the identity preserving loss.  Figure~\ref{fig:pcp} illustrates this procedure with a toy example. 
As illustrated, $H\times W$ number of sub-images $\{x_1, x_2, \dots, x_{HW}\}$ in the dimension of $r\times r$ contain different parts of the original image $X$ in the dimension of $rH\times rW$. Though these sub-images are significantly different in appearance (no overlapping pixels), they share the same identity property, \ie, each sub-image can denote the same waifu character as the original image denotes.
We empirically find that the contextual distance built upon these sub-images helps GENs to learn to better preserve the identity information.

Different from the internal patch~\cite{zoran2011learning} statistics with sliding windows or contextual feature points~\cite{mechrez_maintaining_2018} with pre-trained convolution layers, each sub-image in our distance measurement is extracted with a defined operation $\mathcal{P}\mathcal{K}$.
For an image of size $rH\times rW\times C$, $\mathcal{P}\mathcal{K}$ unshuffles (\ie, a twice-inverse transformation of pixel shuffling~\cite{shi_real-time_2016}, see the supplement for code reference) the image into $H\times W$ number of images in a dimension of $r \times r \times C$ with rate $r$, which can be mathematically denoted as
\begin{equation}
  \mathcal{P}\mathcal{K}(I)_{x*y,r_1,r_2,c} = I_{\lfloor x * r_1\rfloor, \lfloor y * r_2\rfloor, c}.
\end{equation}
Each sub-image is then flattened into a vector, which can be seen as a high-dimensional identity representations of the original image.
We can then estimate the contextual distance between the two representations in a global context.

For contextual distance, we follow Mechrez et al.~\cite{mechrez_contextual_2018} and formally define it on two collections of sub-images $\mathcal{P}\mathcal{K}(X)=\{x_i\}$ and $\mathcal{P}\mathcal{K}(Y)=\{y_j\}$, \eg, restored results and ground truths, as follows
\begin{equation}
  \label{eq:cx}
  \mathrm{SPCX}(\mathcal{P}\mathcal{K}(X), \mathcal{P}\mathcal{K}(Y)) = -\frac{1}{N} \sum_{i} \log \sum_{j} A_{i j},
\end{equation}
where $A_{ij}$ can be seen as the second log term of a multivariate kernel density estimation kernel.
In practice, $A_{ij}$ is implemented to be close to a delta function with normalized cosine distance $\tilde{d}(x_i, y_i)$ and bandwidth $h$ as
\begin{equation}
  \label{eq:kernel}
    A_{i j}=\frac{\exp \left(1-\tilde{d}_{i j} / h\right)}{\sum_{l} \exp \left(1-\tilde{d}_{i l} / h\right)}= \begin{cases}\approx 1 & \text { if } \tilde{d}_{i j} \ll \tilde{d}_{i l} ~\forall l \neq j \\ \approx 0 & \text { otherwise }\end{cases}.
\end{equation}
Minimizing such a distance approximates minimizing the divergence of $KL(\mathbb{P}_X \| \mathbb{P}_Y)$ according to Mechrez et al.~\cite{mechrez_maintaining_2018}, and it is similar to the term of adversarial loss defines in Equation~\eqref{eq:gan}.
In practical implementation, for the face dataset FFHQ in the dimension of $512\times 512$, we empirically find that $r=32$ leads to the best learning performance.
% Perhaps the low-bound dimension of facial identity features is $32\times 32$, and thus operation $\mathcal{P}\mathcal{K}$ at the rate of 32 leads to the best trade between realism and identity.

\subsection{Hierarchical Pseudo Connections}
\label{sec:hpc}
In Section~\ref{sec:pcp} we discussed the spatial periodic contextual distance between two images for considering their identity difference.
However, the existing multiple ground truths issue mentioned in Equation~\eqref{eq:avg} still affects the identity preserving capability of the network, although it can learn with a better objective.
Therefore, in this section, we leveraged the hierarchical generation property of GANs, \eg, StyleGAN~\cite{karras_style-based_2019}, which allows high-resolution photo-realistic images to be generated by combining multi-scale features.
By delicately organizing the features in different scales, we enable well-trained GANs to produce multiple results in a single forward pass.
Thus, with more samples in the same identity but different appearances, our proposed spatial periodic contextual distance can better estimate the distributions and their difference, without network architecture modification on the original embedded GANs.

\begin{figure}[h]
  \center
  \includegraphics[width=1\linewidth]{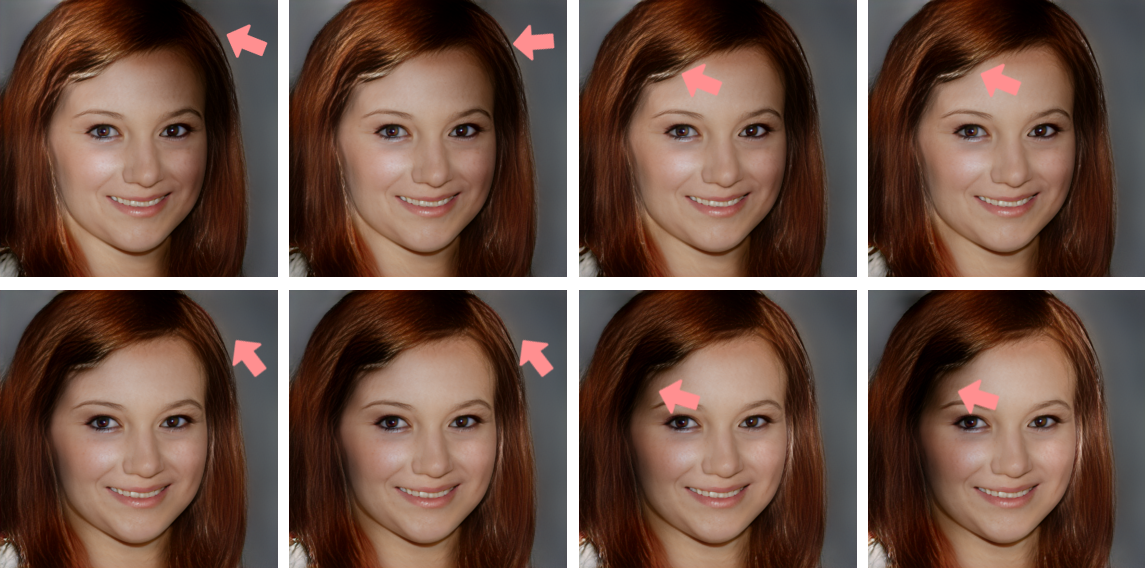}
  \vspace{-1.5\baselineskip}
  \caption{Turbulence degraded images with the same identity but different coarse appearance are restored with HPC $g=3$.
  Here the lighting in the hair part is changing in each image.}
  \label{fig:hpcexp}
  \vspace{-1\baselineskip}
\end{figure}

As Figure~\ref{fig:pipe} illustrated, based on the network architecture of GFPGAN~\cite{wang_towards_2021}, the major difference of our method comes from the proposed \emph{Hierarchical Pseudo Connections (\textbf{HPC})}, which connects the latent conde predictor ($F_{\mathrm{Encoder}}$ and $F_{\mathrm{Decoder}}$) with multiple style blocks for feature modulation.
Here feature modulation~\cite{karras_style-based_2019} is applied as the expanded latent code space for GAN inversion. Our $g$-th HPC transforms feature $M$ from $F_{\mathrm{Decoder}}$ into affine transformation parameters $\{\alpha, \beta\}$, then each group of affine transformation parameters modulates the feature of StyleGAN as
\begin{equation}
  \begin{split}
    \{\alpha_1, \beta_1, \alpha_2, \beta_2, \dots, \alpha_{2^g}, \beta_{2^g} \} = F_{\mathrm{HPC}}^g(M) \\
    F_{\mathrm{StyleGAN}}^i = \alpha_i \odot F_{\mathrm{StyleGAN}} + \beta_i, i \in 2^g.
  \end{split}
\end{equation}
Here we expand each style layer in the original embedded StyleGAN into a group, \ie, grouping style layers with the same parameters but different modulation features.
In Figure~\ref{fig:pipe}, the actual style layer is illustrated with solid lines, and the pseudo style layer is illustrated in dotted lines.
The pseudo style layer is defined as the real style layer processes feature with different modulation features instead of the original modulation features.
Benefited by the hierarchical generation manner, HPC allows the well-trained GAN to generate multiple possible results with similar coarse appearance but different fine details.
Therefore, the applied HPC with $2^g$ number of style layers in the final group, comes from groups consisting of $\{2^1, 2^2, \dots, 2^{g-1}\}$ number of style layers respectively, which transforms $G(\cdot)$ as 
\begin{equation}
  \mathbf{\hat{I}} = \{\hat{I}_1, \hat{I}_2, \dots, \hat{I}_{2^g}\} = G(\tilde{I}).
\end{equation}
Assume that the ill-posed nature of problem produces $n$ possible solutions in the image set $\mathbf{I}$.
Though $2^g << n$, we can significantly reduces their difference with the transformed Equation~\eqref{eq:cx} as
\begin{equation}
  \label{eq:vgan}
  \mathcal{L}(G) = SPCX(\mathcal{P}\mathcal{K}(I)), \mathcal{P}\mathcal{K}(\mathbf{\hat{I}}), r).
\end{equation}
We then stack all of the generated results for training the network and take the averaged image as the restoration result.
A generated example from our method with $g=3$ is shown in Figure~\ref{fig:hpcexp}, where 8 similar images are generated.

\subsection{Model Objective}
The final learning objective for training our method combines both the proposed spatial periodic contextual loss, adversarial loss, perceptual loss, and identity preserving loss.

\
\noindent \textbf{Reconstruction Related.}
For network $G(*)$ with $2^g$ number of final hierarchical pseudo connections we have
\begin{equation}
  \begin{split}
  \mathcal{L}_{rec} (G) & = \frac{1}{2^g} \sum_i^{2^g} SPCX(\mathcal{P}\mathcal{K}(I),\mathcal{P}\mathcal{K}(G(\tilde{I})_i)) \\
  & - \lambda_{adv} \mathbb{E}_{G(\tilde{I})_i} \operatorname{softplus}(D(G(\tilde{I})_i)) \\
  & + \lambda_{per} \|\phi(I) - \phi(G(\tilde{I})_i) \| \\
  & + \lambda_{id} \|\eta(I) - \eta(G(\tilde{I})_i) \|,
  \end{split}
\end{equation}
where $\phi(*)$ is the pre-trained VGG-19 network~\cite{simonyan2014very} that is applied with its $\{\mathrm{conv}1,\dots,\mathrm{conv}5 \}$ layers before activation~\cite{wang2018esrgan}, $\eta(*)$ is the pre-trained ArcFace network~\cite{deng_arcface_2019} without the final logistic layer.
$\lambda_{adv}$, $\lambda_{per}$, and $\lambda_{id}$ are the loss weights of adversarial loss, perceptual loss, and identity preserving loss, respectively.
In our implementation, we empirically set them as $\{1:0.1:10\}$.

\noindent \textbf{Adversarial Related.}
The discriminator $D(*)$ is trained similar to StyleGAN2~\cite{karras_analyzing_2020} except that multiple results are applied and the adversarial losses are averaged.

\section{Experiments}
\label{sec:exp}
In this section, we present the experimental details for evaluating our method and its settings, as well as the comparison results with the state-of-the-art methods.

\subsection{Testing and Training Settings}
\noindent \textbf{Synthesized Testing Benchmark.}
For the reference-based evaluation, we synthesize turbulence-clean image pairs using TurbulenceSim\_P2S~\cite{mao2021accelerating} which is the current state-of-the-art turbulence simulation works.
The synthesis is conducted on the selected first 100 images of CelebAHQ~\cite{karras_progressive_2017}, named \textbf{CelebAHQ100}, and the parameters of TurbulenceSim\_P2S are carefully selected to match the real-world turbulence images.
Specifically, we set \emph{D}, \emph{r0}, and \emph{corr} as $\{5, 1.25, -0.01\}$ for the CelebAHQ100 simulation.
The synthesized image pairs are provided in the supplementary document in 8-bit sRGB format with a resolution of $512\times 512$.\\

\begin{figure}[htbp]
  \begin{subfigure}[t]{0.24\linewidth}
    \captionsetup{justification=centering, labelformat=empty, font=scriptsize}
    \includegraphics[width=1\linewidth]{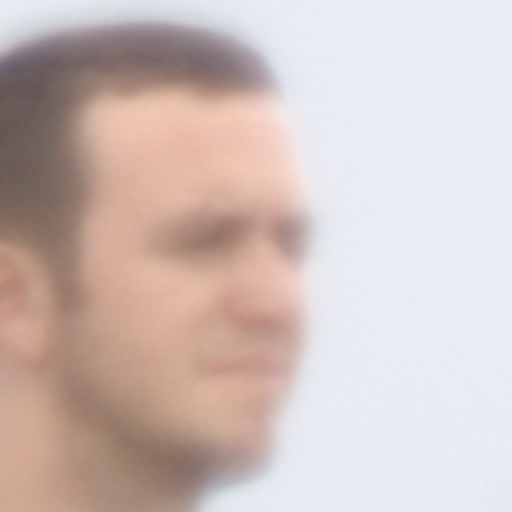}
    \caption{Left Position}
  \end{subfigure}
  \begin{subfigure}[t]{0.24\linewidth}
    \captionsetup{justification=centering, labelformat=empty, font=scriptsize}
    \includegraphics[width=1\linewidth]{figures/teaser/tubimages/019.png}
    \caption{Center Position}
  \end{subfigure}
  \begin{subfigure}[t]{0.24\linewidth}
    \captionsetup{justification=centering, labelformat=empty, font=scriptsize}
    \includegraphics[width=1\linewidth]{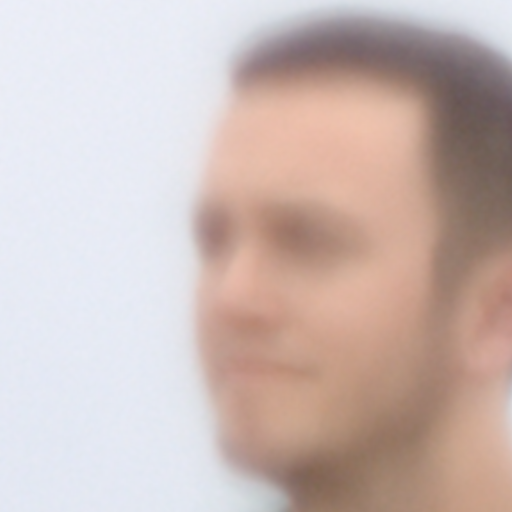}
    \caption{Right Position}
  \end{subfigure}
  \begin{subfigure}[t]{0.24\linewidth}
    \captionsetup{justification=centering, labelformat=empty, font=scriptsize}
    \includegraphics[width=1\linewidth]{figures/teaser/references/07.png}
    \caption{Indoor Reference}
  \end{subfigure}
  \vspace{-0.5\baselineskip}
  \caption{Visualization of applied real-world turbulence images. These images are captured by cameras put in left, center, right positions, and the last image is captured indoor without turbulence.}
  \vspace{-1\baselineskip}
  \label{fig:sample}
\end{figure}

\vspace{-0.5\baselineskip}
\noindent \textbf{Real-world Testing Benchmark.} 
For evaluating the performance of different methods on real-world turbulence degraded images without pixel-wise corresponding ground truths, we use face recognition accuracy based on indoor reference clear images.
The authors of~\cite{yasarla_learning_2020} provided us high-quality raw real-world turbulence degraded faces which are taken at 300 meters from the camera in a hot day.  In addition to those images, we received the corresponding indoor face images without turbulence for reference.
We crop and wrap faces with the pre-trained RetinaFace~\cite{deng_retinaface_2019} network.
The final dataset contains images from 89 separate individuals each having 3 turbulence degraded images in different poses.  We call this data \textbf{TubFace89} and show its sampled images in Figure~\ref{fig:sample}.

\begin{figure*}[htbp]
  \begin{subfigure}[t]{.12\linewidth}
    \captionsetup{justification=centering, labelformat=empty, font=scriptsize}
    \includegraphics[width=1\linewidth]{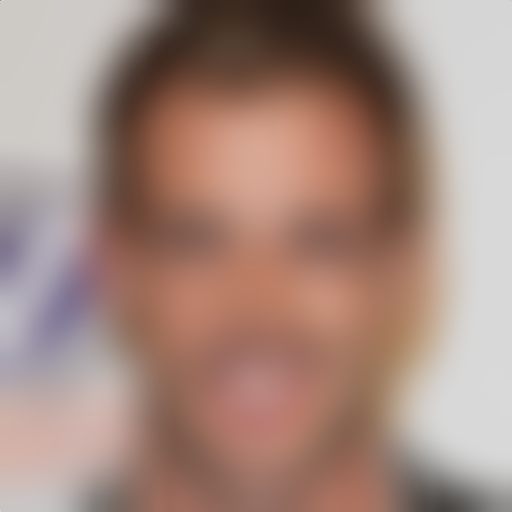}
    \includegraphics[width=1\linewidth]{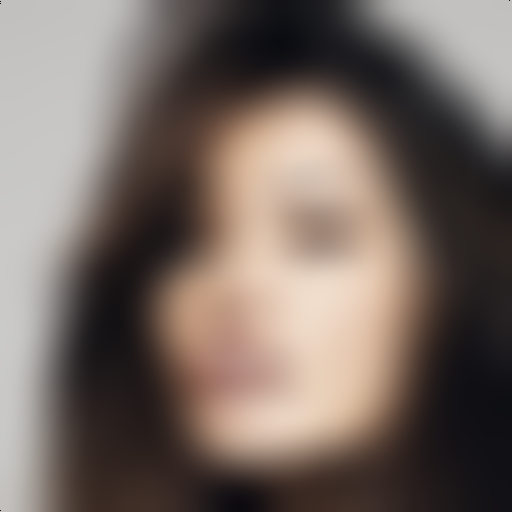}
    \includegraphics[width=1\linewidth]{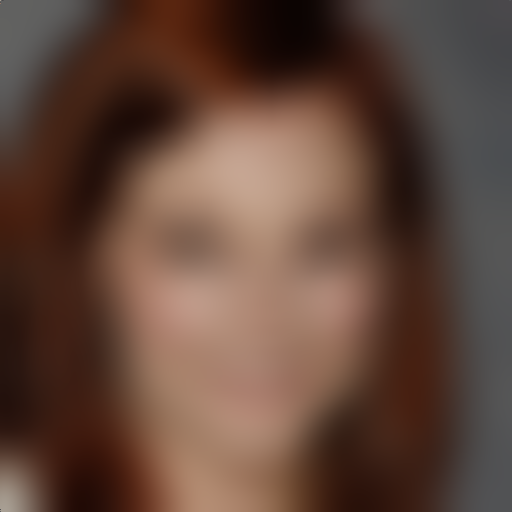}
    \caption{Turbulence Image}
  \end{subfigure}
  \begin{subfigure}[t]{.12\linewidth}
    \captionsetup{justification=centering, labelformat=empty, font=scriptsize}
    \includegraphics[width=1\linewidth]{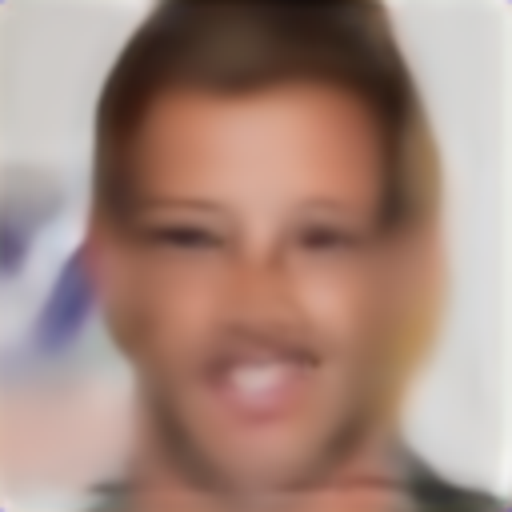}
    \includegraphics[width=1\linewidth]{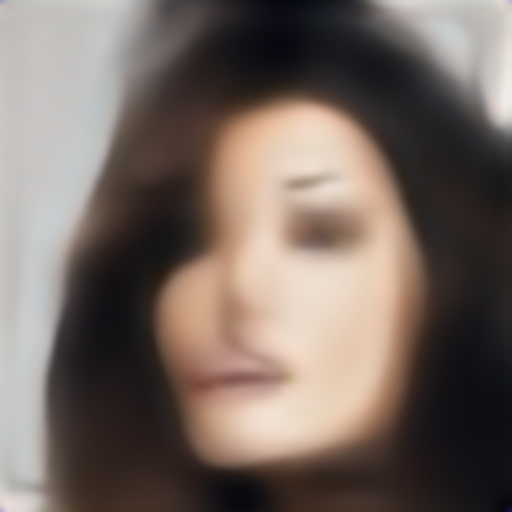}
    \includegraphics[width=1\linewidth]{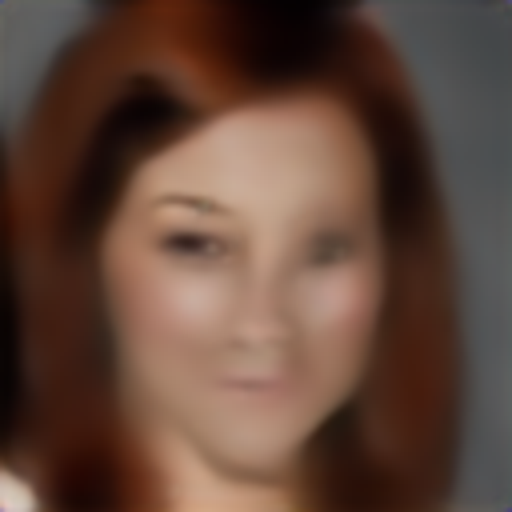}
    \caption{TDRN \\ \emph{arXiv20}}
  \end{subfigure}
  \begin{subfigure}[t]{.12\linewidth}
    \captionsetup{justification=centering, labelformat=empty, font=scriptsize}
    \includegraphics[width=1\linewidth]{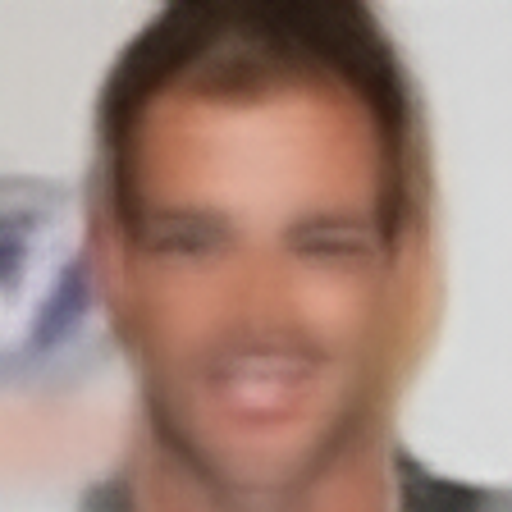}
    \includegraphics[width=1\linewidth]{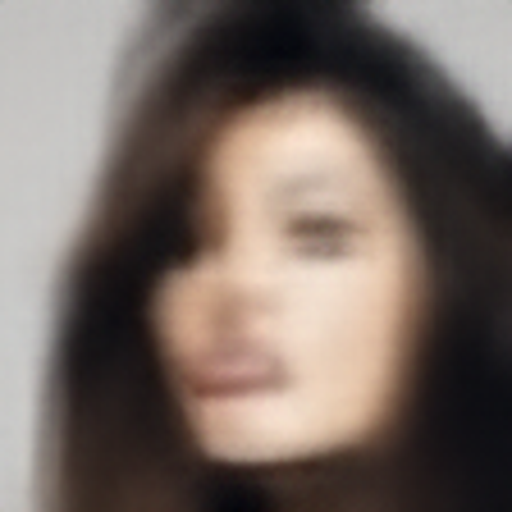}
    \includegraphics[width=1\linewidth]{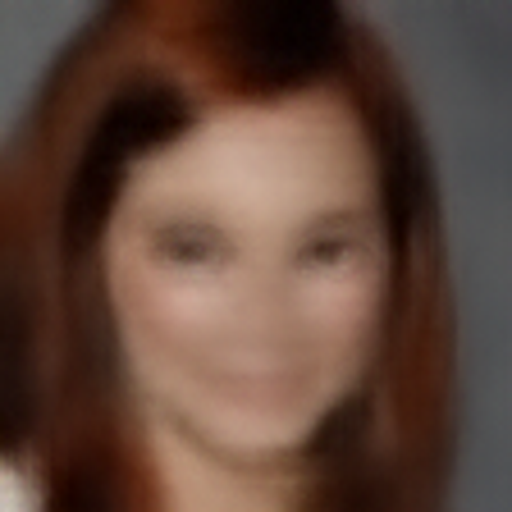}
    \caption{ATFaceGAN \\ \emph{TBIOM21}}
  \end{subfigure}
  \begin{subfigure}[t]{.12\linewidth}
    \captionsetup{justification=centering, labelformat=empty, font=scriptsize}
    \includegraphics[width=1\linewidth]{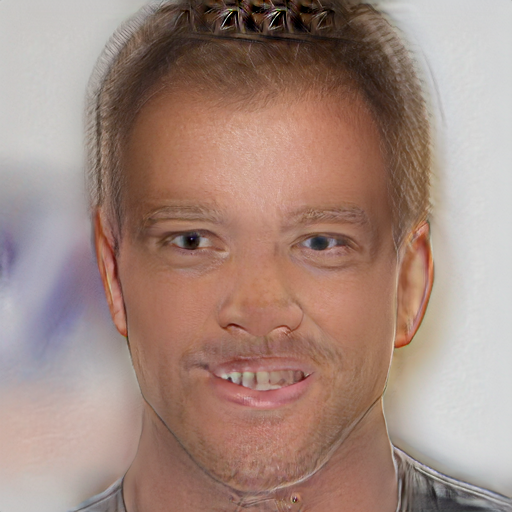}
    \includegraphics[width=1\linewidth]{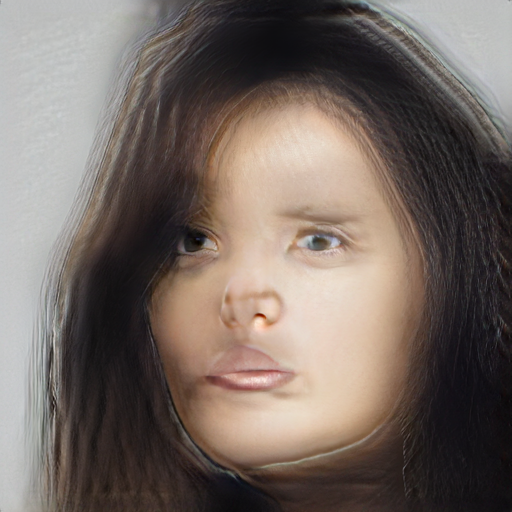}
    \includegraphics[width=1\linewidth]{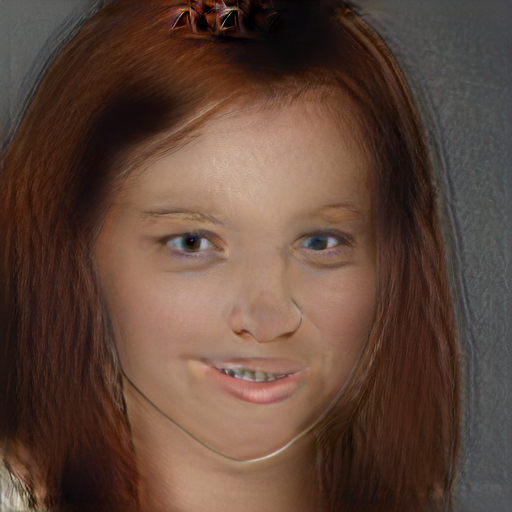}
    \caption{PSFRGAN \\ \emph{CVPR21}}
  \end{subfigure}
  \begin{subfigure}[t]{.12\linewidth}
    \captionsetup{justification=centering, labelformat=empty, font=scriptsize}
    \includegraphics[width=1\linewidth]{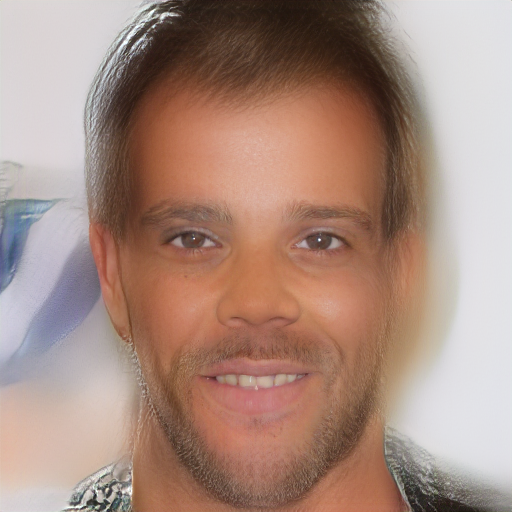}
    \includegraphics[width=1\linewidth]{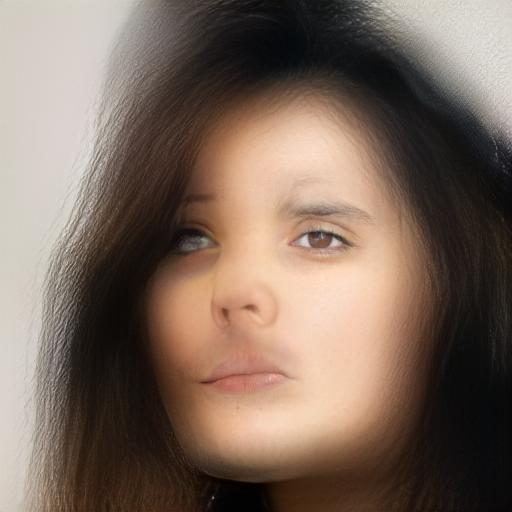}
    \includegraphics[width=1\linewidth]{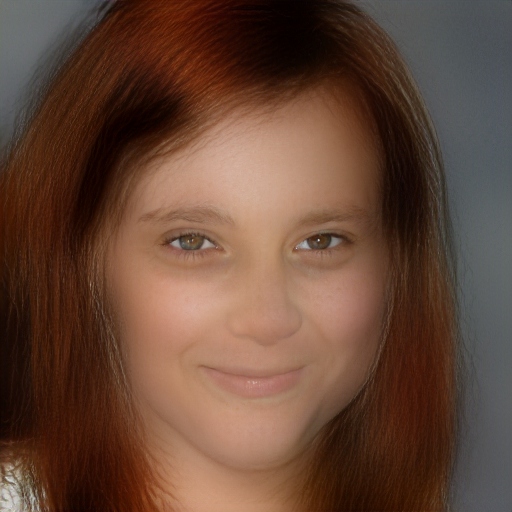}
    \caption{GFPGAN \\ \emph{CVPR21}}
  \end{subfigure}
  \begin{subfigure}[t]{.12\linewidth}
    \captionsetup{justification=centering, labelformat=empty, font=scriptsize}
    \includegraphics[width=1\linewidth]{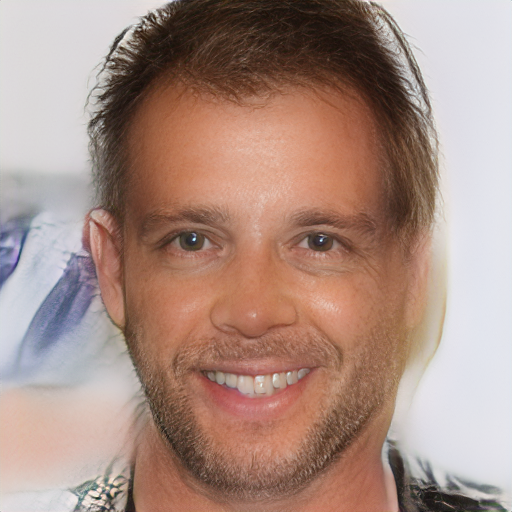}
    \includegraphics[width=1\linewidth]{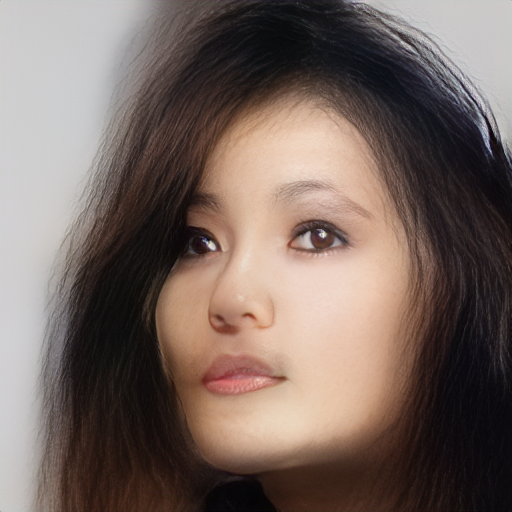}
    \includegraphics[width=1\linewidth]{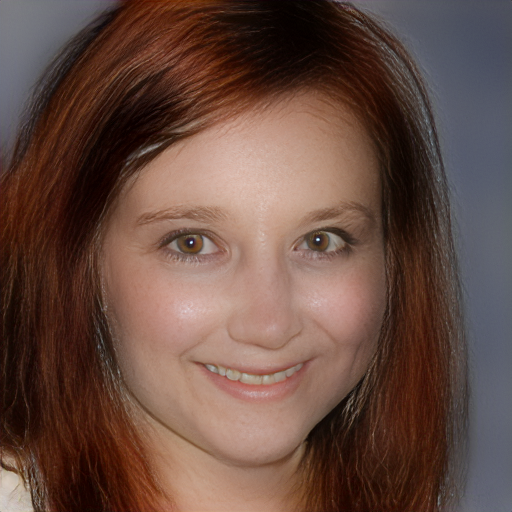}
    \caption{GFPGAN* \\ (ElasticAug)}
  \end{subfigure}
  \begin{subfigure}[t]{.12\linewidth}
    \captionsetup{justification=centering, labelformat=empty, font=scriptsize}
    \includegraphics[width=1\linewidth]{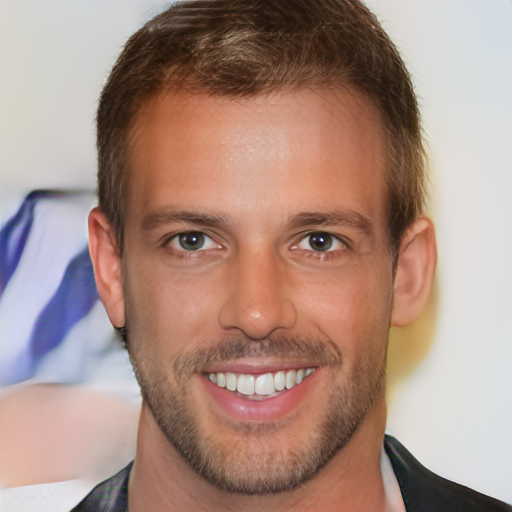}
    \includegraphics[width=1\linewidth]{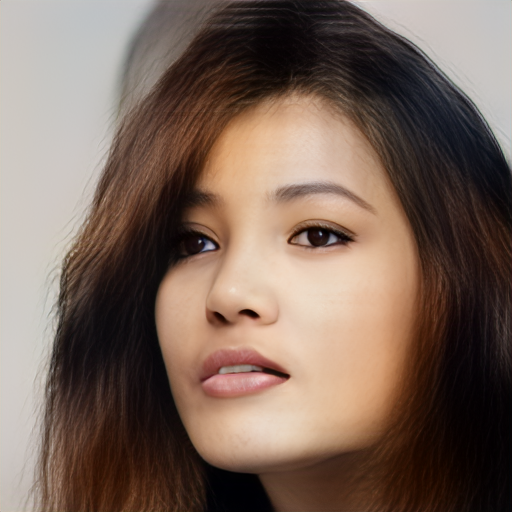}
    \includegraphics[width=1\linewidth]{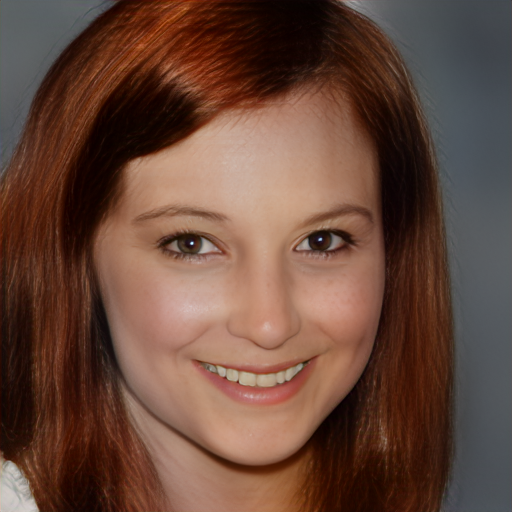}
    \caption{\textbf{LTT-GAN \\ \emph{Ours}}}
  \end{subfigure}
  \begin{subfigure}[t]{.12\linewidth}
    \captionsetup{justification=centering, labelformat=empty, font=scriptsize}
    \includegraphics[width=1\linewidth]{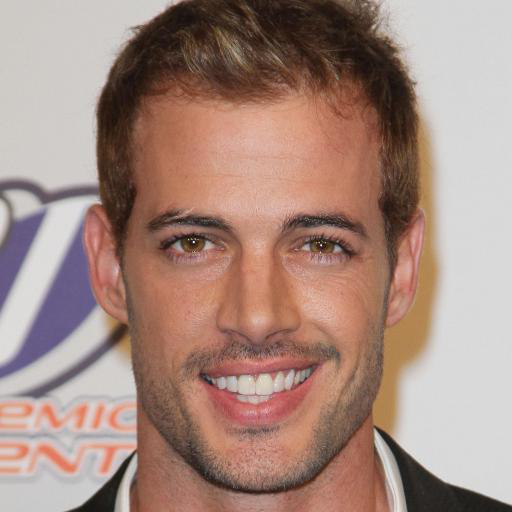}
    \includegraphics[width=1\linewidth]{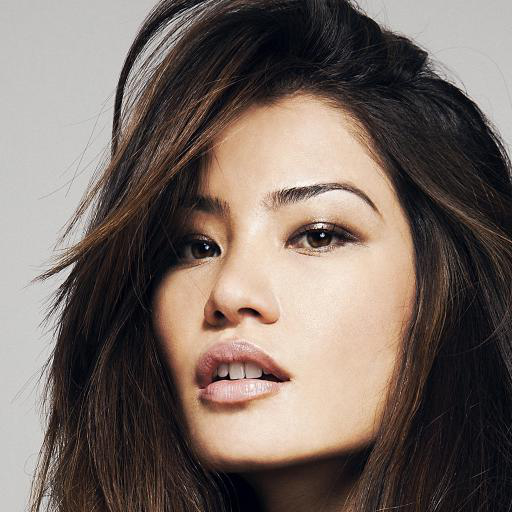}
    \includegraphics[width=1\linewidth]{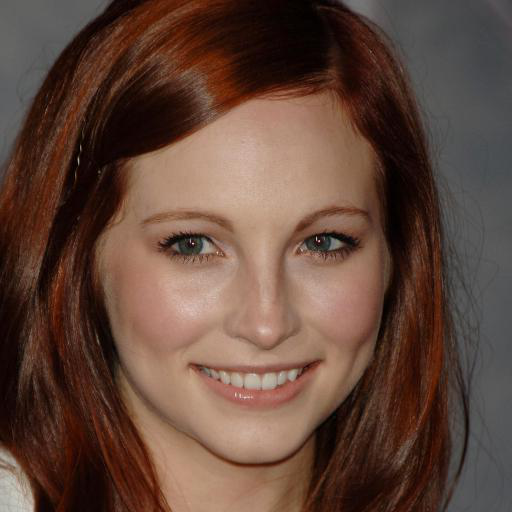}
    \caption{Ground Truth \\ Image}
  \end{subfigure}
  \hfill
  \vspace{-0.5\baselineskip}
  \caption{Visualization results of restored synthesized turbulence face images. Compared with results contain unnature artifacts, our results achieve the best visual quality and identity similarity of the ground truth. See the supplement for more such comparisons.}
  \label{fig:syncomp}
  \vspace{-1\baselineskip}
\end{figure*}

\begin{figure}[htbp]
  \centering
  \begin{subfigure}[t]{0.24\linewidth}
    \captionsetup{justification=centering, labelformat=empty, font=scriptsize}
    \includegraphics[width=1\linewidth]{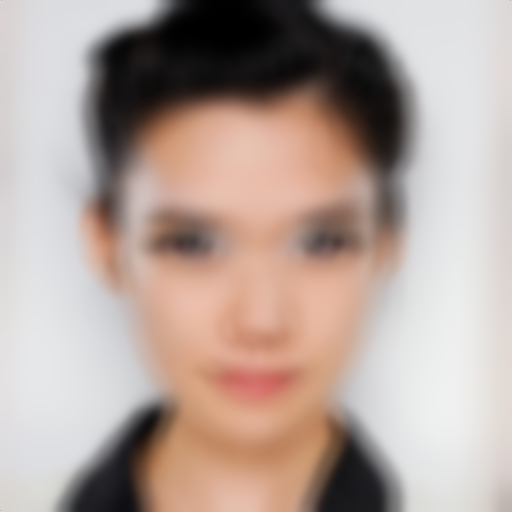}
    \caption{Synthesized Input}
  \end{subfigure}
  \begin{subfigure}[t]{0.24\linewidth}
    \captionsetup{justification=centering, labelformat=empty, font=scriptsize}
    \includegraphics[width=1\linewidth]{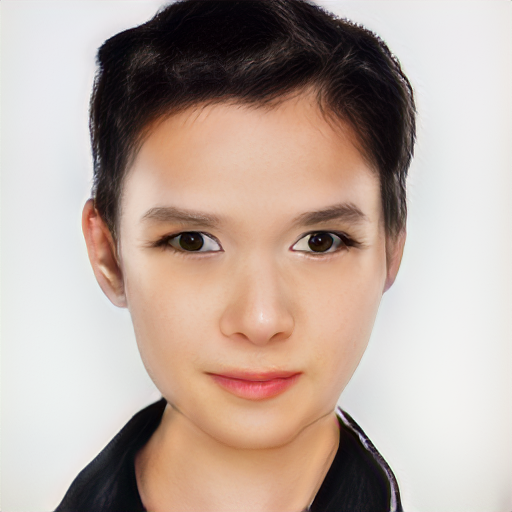}
    \caption{BFR}
  \end{subfigure}
  \begin{subfigure}[t]{0.24\linewidth}
    \captionsetup{justification=centering, labelformat=empty, font=scriptsize}
    \includegraphics[width=1\linewidth]{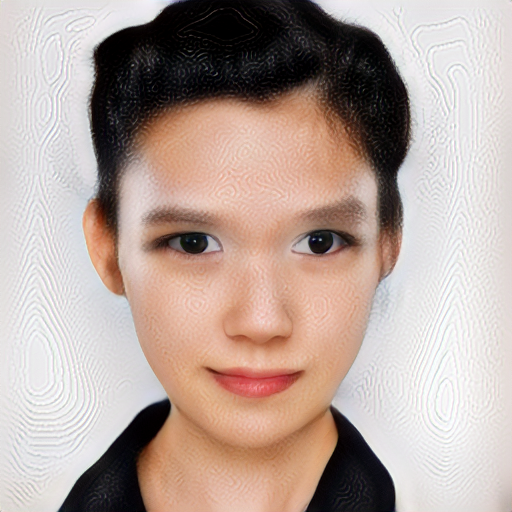}
    \caption{TurbulenceSim\_P2S}
  \end{subfigure}
  \begin{subfigure}[t]{0.24\linewidth}
    \captionsetup{justification=centering, labelformat=empty, font=scriptsize}
    \includegraphics[width=1\linewidth]{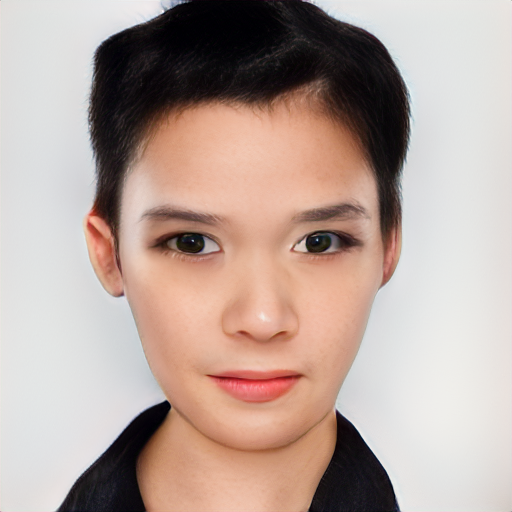}
    \caption{Ours}
  \end{subfigure}
  \hfill
  \vspace{-0.5\baselineskip}
  \caption{Contour and ring artifacts are shown in the restored results of networks trained on simulated images from TurbulenceSim and BFR. In contrast, the result from the network trained on our ElasticAug shows sharper edges and more accurate facial details. \textbf{(200\% Zoom is recommended to see their difference.)}}
  \label{fig:contour} 
  \vspace{-1.5\baselineskip}
\end{figure}

\vspace{0.5\baselineskip}
\noindent \textbf{Training Dataset and Data Augmentation}
We applied FFHQ~\cite{karras_style-based_2019} with 70,000 high-quality images in a resolution of $512 \times 512$ as the training dataset.
To find the best way to boost the turbulence mitigation capability of networks, we conduct experiments with three existing data augmentation methods, \ie, Blind Face Restoration (BFR)~\cite{li_learning_2018, li_blind_2020,yang_gan_2021,wang_towards_2021}, TurbulenceSim\_P2S~\cite{mao2021accelerating}, and our new method, called ElasticAug for simulation turbulence effects for training.

Among them, TurbulenceSim\_P2S learns the basis functions for spatially varying convolutions from known turbulence models for turbulence simulation.
Since the real-world turbulence images usually suffer strong blur degradation, we empirically find that the degradation procedure used in BFR can also produce similar degraded results to real-world turbulence degradation.
Moreover, in this paper, we introduce a new simulation method, called \textbf{ElasticAug}, which is based on BFR augmentation, and it combines the blur augmentation with Elastic transformation~\cite{simard_best_2003}, which randomly moves pixels using local displacement fields.

Here we directly apply GFPGAN~\cite{wang_towards_2021} as the baseline and train the network on three data augmentation methods.
Figure~\ref{fig:contour} shows their restored results.
One can find that the network trained on ElasticAug achieves the best visual quality of restored results.
Although BFR produces a similar degradation procedure of turbulence, it cannot estimate the spatial deformation, and hence its result contains strong ring artifacts.
Although TurbulenceSim\_P2S produces a comparable clear result, the network trained on its simulated images generates the result with strange contour artifacts, and thus the data augmentation is only used for evaluation.
Therefore, our proposed ElasticAug is availed as the default data augmentation method in the following experiments.

\vspace{0.5\baselineskip}
\noindent \textbf{Implementation} 
Our implementation follows the settings of conventional generative embedding networks~\cite{wang_towards_2021}, \ie, the learning rate was set to $2\times 10^{-3}$ and decayed at 600k and 700k iterations by a rate of 0.5, with a total of 800k iterations.
The whole training is conducted in a mini-batch size of 12 using 4 NVIDIA A100 GPUs.
For the $F_{Encoder}$ and $F_{Decoder}$, we initialized their parameters with Xavier initialization.
For the GAN module, we apply StyleGAN2~\cite{karras_analyzing_2020} with its well-trained parameters on FFHQ generation, and its parameters are frozen at the network training.
For the discriminator module, we apply the discriminator of well-trained StyleGAN2 on FFHQ generation, and its parameters are optimized with the same learning rate as the GAN module. Here we empirically set $r=32$ and $g=3$.

\begin{figure*}[htbp]
  \begin{subfigure}[t]{.12\linewidth}
    \captionsetup{justification=centering, labelformat=empty, font=scriptsize}
    \includegraphics[width=1\linewidth]{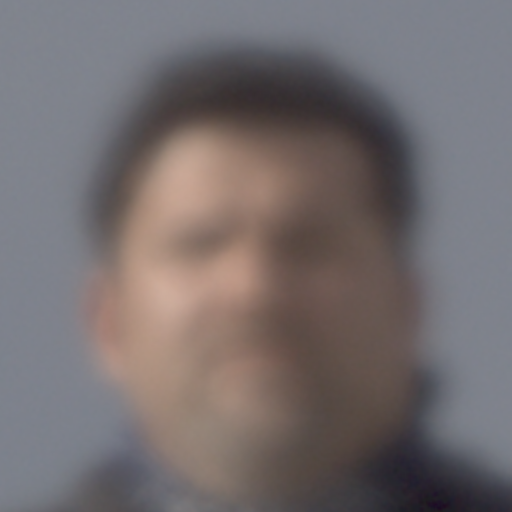}
    \includegraphics[width=1\linewidth]{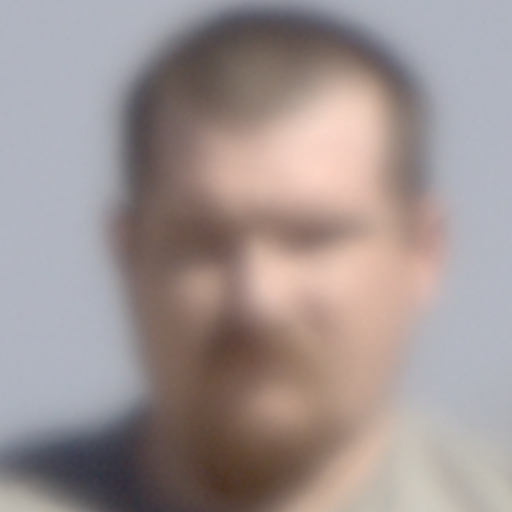}
    \includegraphics[width=1\linewidth]{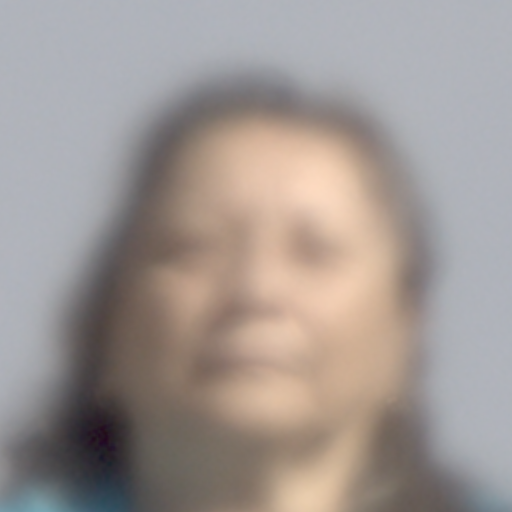}
    \caption{Turbulence \\ Image}
  \end{subfigure}
  \begin{subfigure}[t]{.12\linewidth}
    \captionsetup{justification=centering, labelformat=empty, font=scriptsize}
    \includegraphics[width=1\linewidth]{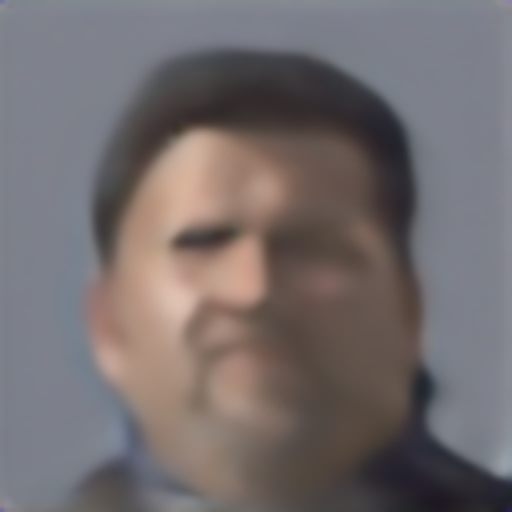}
    \includegraphics[width=1\linewidth]{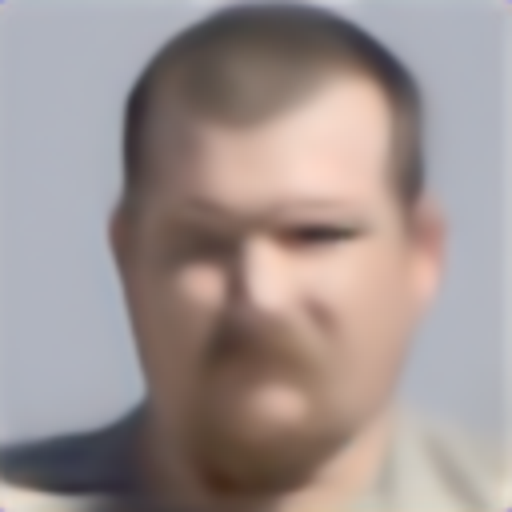}
    \includegraphics[width=1\linewidth]{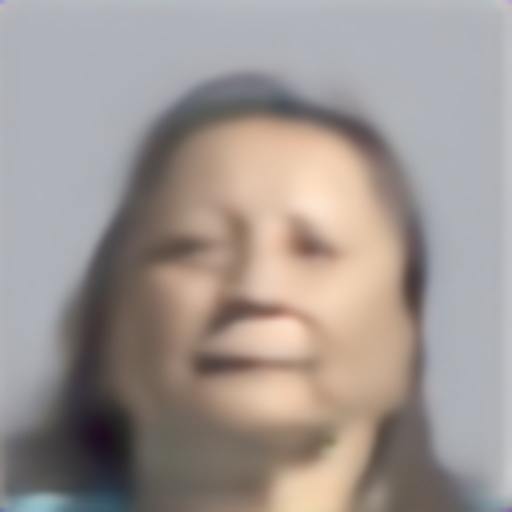}
    \caption{TDRN \\ \emph{arXiv20}}
  \end{subfigure}
  \begin{subfigure}[t]{.12\linewidth}
    \captionsetup{justification=centering, labelformat=empty, font=scriptsize}
    \includegraphics[width=1\linewidth]{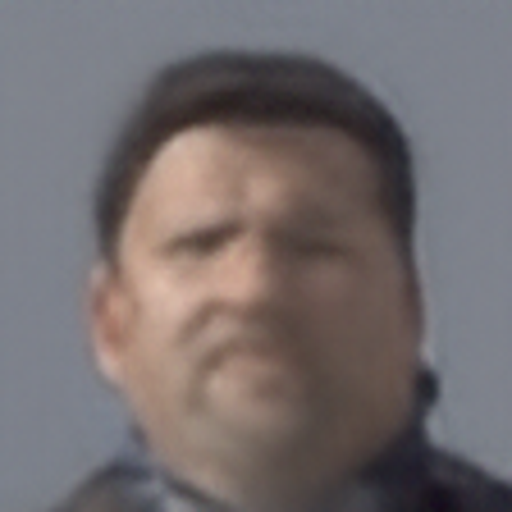}
    \includegraphics[width=1\linewidth]{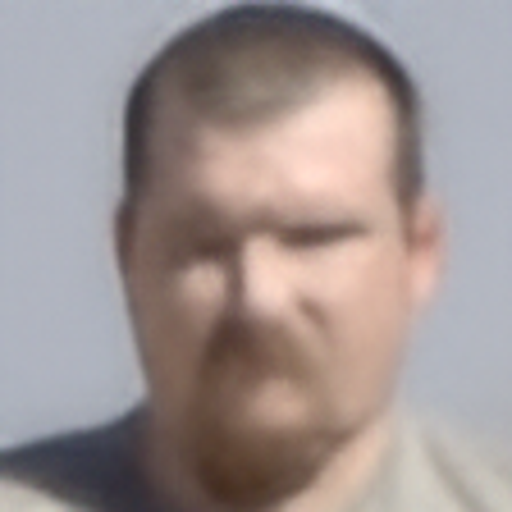}
    \includegraphics[width=1\linewidth]{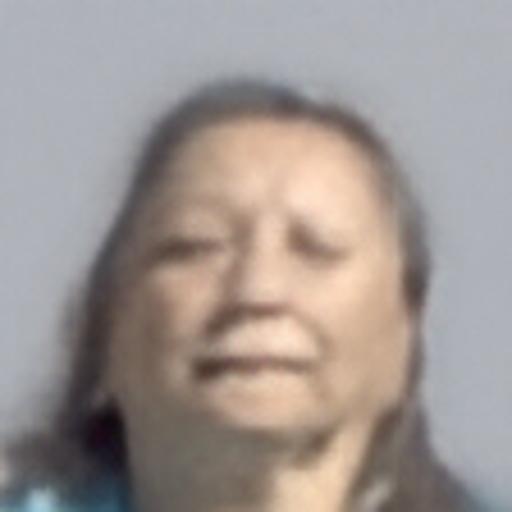}
    \caption{ATFaceGAN \\ \emph{TBIOM21}}
  \end{subfigure}
  \begin{subfigure}[t]{.12\linewidth}
    \captionsetup{justification=centering, labelformat=empty, font=scriptsize}
    \includegraphics[width=1\linewidth]{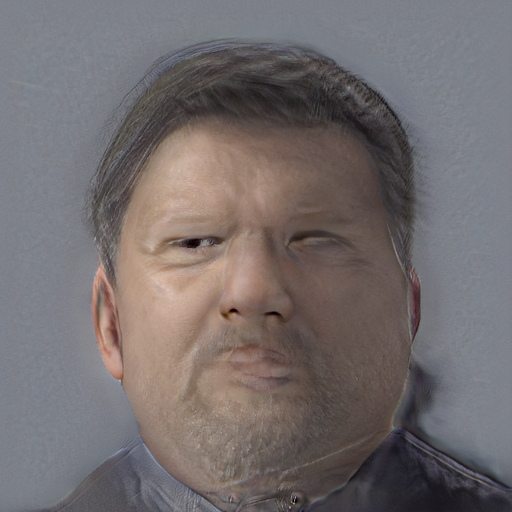}
    \includegraphics[width=1\linewidth]{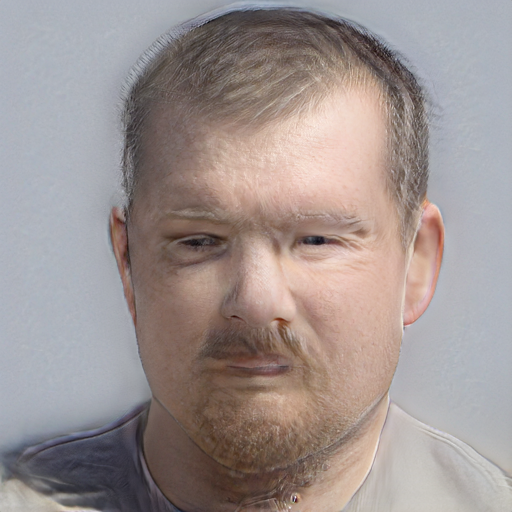}
    \includegraphics[width=1\linewidth]{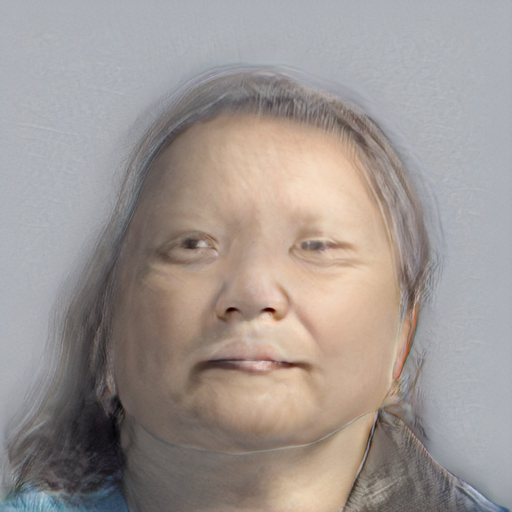}
    \caption{PSFRGAN \\ \emph{CVPR21}}
  \end{subfigure}
  \begin{subfigure}[t]{.12\linewidth}
    \captionsetup{justification=centering, labelformat=empty, font=scriptsize}
    \includegraphics[width=1\linewidth]{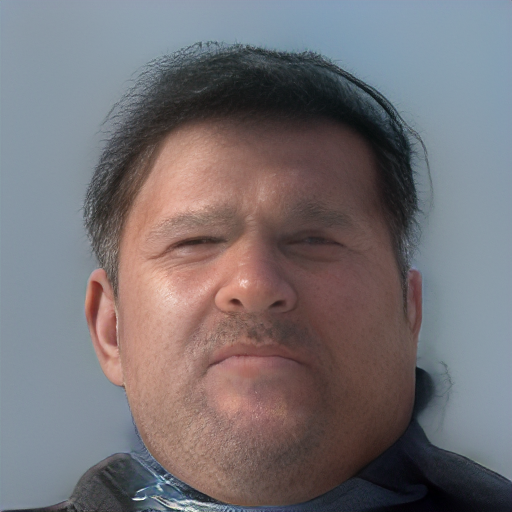}
    \includegraphics[width=1\linewidth]{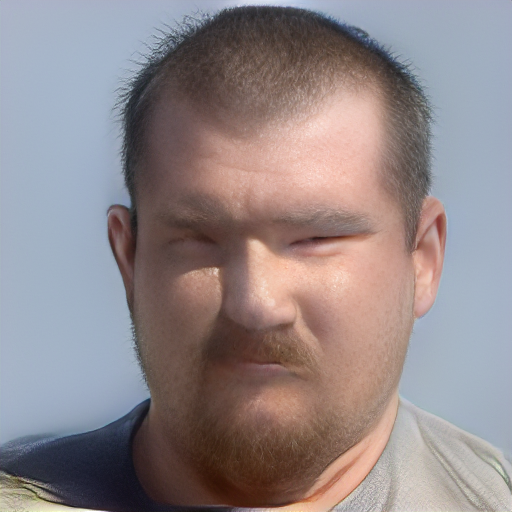}
    \includegraphics[width=1\linewidth]{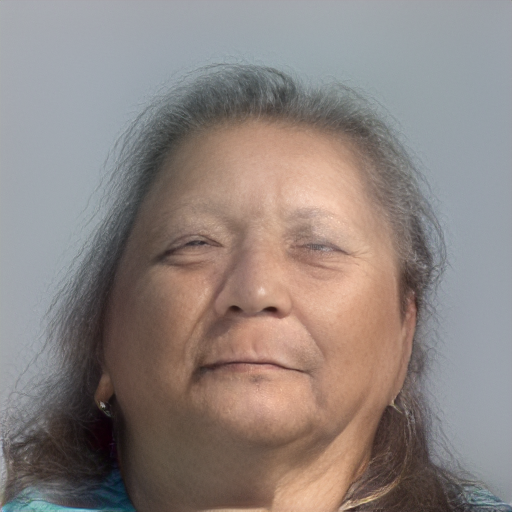}
    \caption{GFPGAN \\ \emph{CVPR21}}
  \end{subfigure}
  \begin{subfigure}[t]{.12\linewidth}
    \captionsetup{justification=centering, labelformat=empty, font=scriptsize}
    \includegraphics[width=1\linewidth]{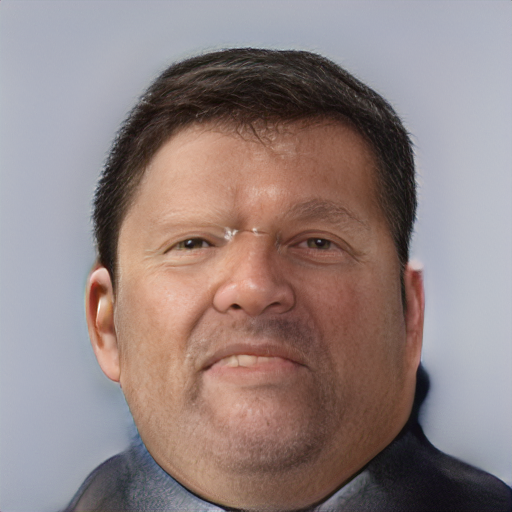}
    \includegraphics[width=1\linewidth]{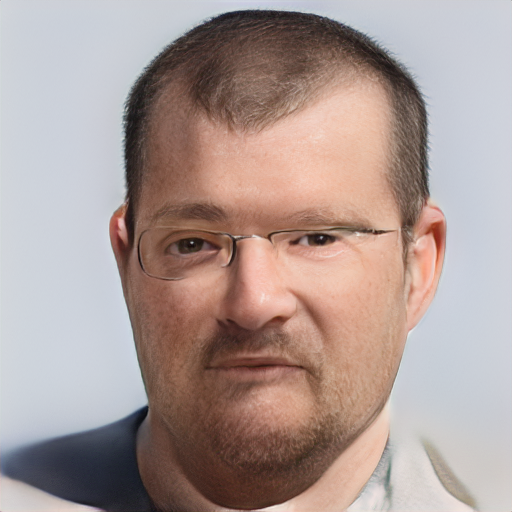}
    \includegraphics[width=1\linewidth]{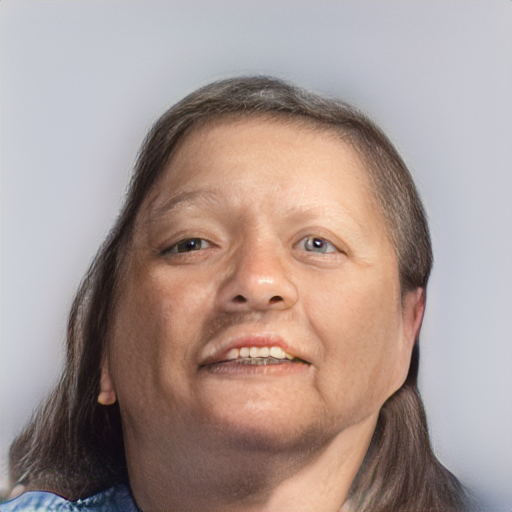}
    \caption{GFPGAN* \\ (ElasticAug)}
  \end{subfigure}
  \begin{subfigure}[t]{.12\linewidth}
    \captionsetup{justification=centering, labelformat=empty, font=scriptsize}
    \includegraphics[width=1\linewidth]{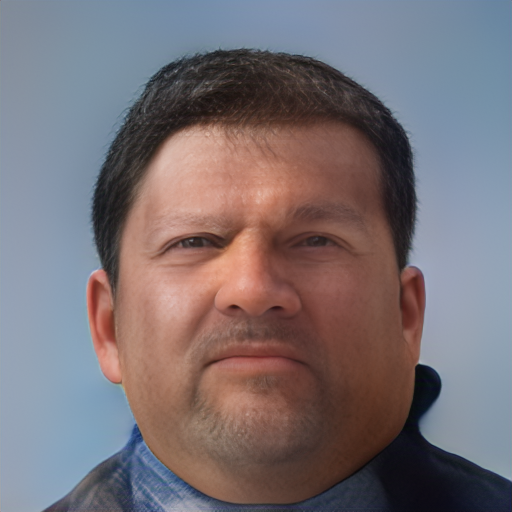}
    \includegraphics[width=1\linewidth]{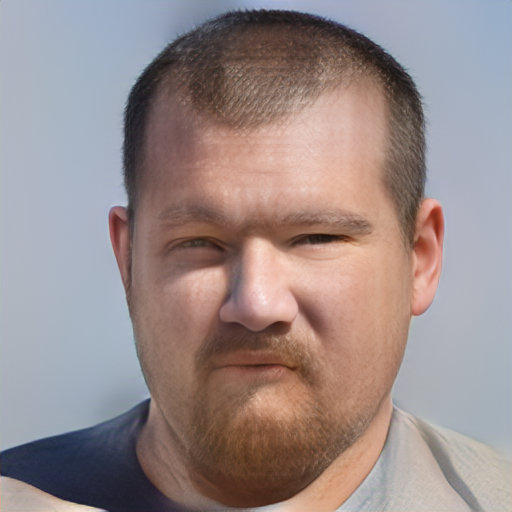}
    \includegraphics[width=1\linewidth]{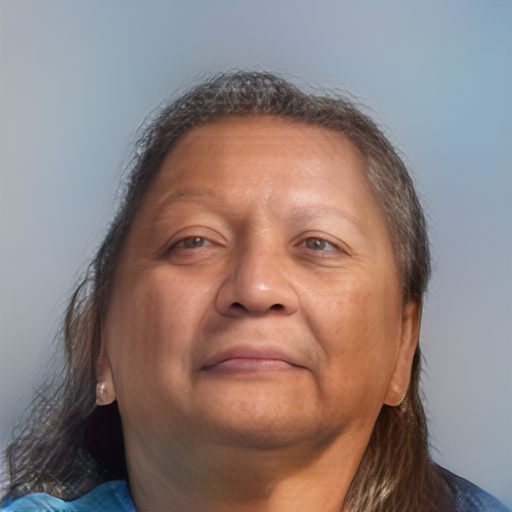}
    \caption{\textbf{LTT-GAN \\ \emph{Ours}}}
  \end{subfigure}
  \begin{subfigure}[t]{.12\linewidth}
    \captionsetup{justification=centering, labelformat=empty, font=scriptsize}
    \includegraphics[width=1\linewidth]{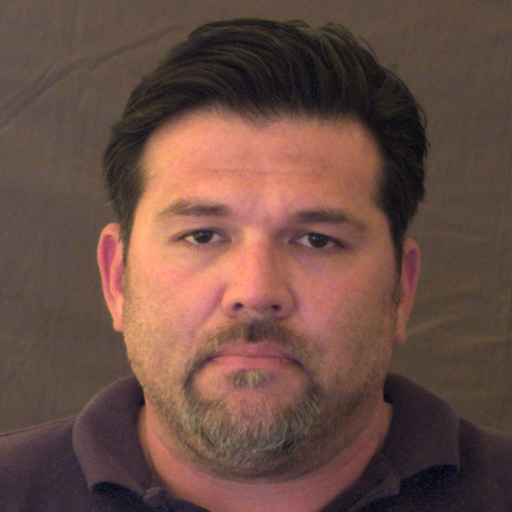}
    \includegraphics[width=1\linewidth]{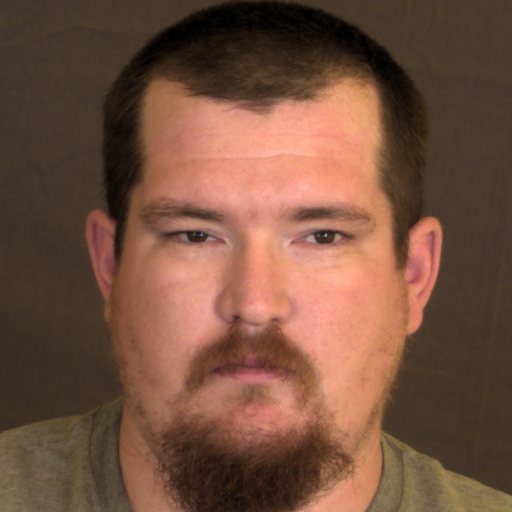}
    \includegraphics[width=1\linewidth]{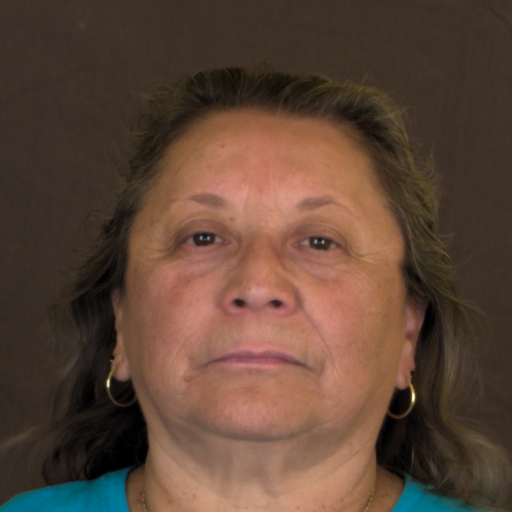}
    \caption{Ground Truth \\ Image}
  \end{subfigure}
  \hfill
  \vspace{-0.5\baselineskip}
  \caption{Visualization results of compared methods on real-world turbulence images. The proposed method achieves the best visual quality and maintains the identity in the restored images with fewer unfaithful details. See the supplement for more such comparisons.}
  \label{fig:realcomp}
  \vspace{-0.5\baselineskip}
\end{figure*}

\begin{table*}[htbp]
  \centering
  \resizebox{\linewidth}{!}{
  \begin{tabular}{@{}lcccccccccc@{}}
    \toprule
    & \!BFR\! & \!ElasticAug\! & \!LPIPS $\downarrow$\! & \!FID $\downarrow$\! & \!NIQE $\downarrow$\! & \!Deg. (\%) $\uparrow$\! & \!PSNR $\uparrow$\! & \!SSIM $\uparrow$\! & \! Imgs/Sec \! & \! \ Params (M) \!  \\
    \midrule
    Turbulence Images & - & - & 0.6490 & 274.84 & 17.48 & 8.91 & 20.33 & \cellcolor{yellow} 0.6455 & - & - \\
    PSFRGAN~\cite{chen_progressive_2021} [CVPR21] & \checkmark & - & 0.4070 & 122.39 & \cellcolor{red} 4.051 & 22.30 & 19.96 & 0.5451 & 3.70 & 184.2 \\
    GFPGAN~\cite{wang_towards_2021} [CVPR21] & \checkmark & - & \cellcolor{yellow} 0.3800 & \cellcolor{yellow} 113.42 & 5.515 & \cellcolor{yellow} 26.55 & 20.31 & 0.5797 & 20.14 & 615.4  \\
    TDRN~\cite{yasarla_learning_2020} [arXiv20] & - & \checkmark & 0.5869 & 190.17 & 13.55 & 18.54 & \cellcolor{orange} 21.63 & \cellcolor{orange} 0.6611 & 3.41 & 8 \\
    ATFaceGAN~\cite{lau_atfacegan_2021} [TBIOM21] & - & \checkmark & 0.5868 & 181.10 & 12.71 & 15.55 & \cellcolor{red} 21.77 & \cellcolor{red} 0.6633 & 7.32 &  68.70 \\
    GFPGAN* & - & \checkmark & \cellcolor{orange} 0.3288 & \cellcolor{orange} 90.23 & \cellcolor{yellow} 4.435 & \cellcolor{orange} 40.53 & 20.49 & 0.5790 & 20.14 & 615.4  \\
    LTT-GAN (Ours) & - & \checkmark & \cellcolor{red} 0.2906 & \cellcolor{red} 85.72 & \cellcolor{orange} 4.285 & \cellcolor{red} 49.88 & 20.96 & \cellcolor{yellow} 0.6042 &  14.12 & 741.4 \\
    \bottomrule
  \end{tabular}
  }
  \vspace{-0.5\baselineskip}
  \caption{Performance comparisons against the state-of-the-art methods and ours on the synthesized turbulence face images.  achieves the best performance on visual quality, perceptual metrics, and identity-preserving metrics, \ie, FID, LPIPS, and Deg.}
  \label{tab:syncomp}
  \vspace{-1\baselineskip}
\end{table*}

\subsection{Comparisons with SOTA Methods}
We conduct comparisons with several state-of-the-art turbulence mitigation methods \ie, TDRN~\cite{yasarla_learning_2020} and ATFaceGAN~\cite{lau_atfacegan_2021}.
For fair comparisons, we finetune them on FFHQ with their officially released codes.
We also compare with the recent SOTA blind face restoration methods, \ie, PSFRGAN~\cite{chen_progressive_2021} and GFPGAN~\cite{wang_towards_2021} trained on FFHQ. 

\vspace{0.5\baselineskip}
\noindent \textbf{Synthesized Turbulence.}  In Figure~\ref{fig:syncomp}, we compare visual restored results on the synthesized turbulence degraded images.
One can find that our method best restores the facial details with photorealism.
Among the compared methods, though the results from GFPGAN trained on ElasticAug seem to contain comparable details, their details are less similar to the ground truth. In comparison, our results best preserve the identity related details in the restored images.

In Table~\ref{tab:syncomp}, we present quantitative performance comparisons in the widely used visual quality metrics, \ie, LPIPS~\cite{zhang_unreasonable_2018}, FID~\cite{heusel_gans_2017}, NIQE~\cite{mittal_making_2012}, and pixel-wise metrics, \ie, PSNR and SSIM.
Since the turbulence mitigation task is highly correlated to face recognition, we further employ the identity metric, \ie, Deg, which is the cosine distance of facial features on the pretrained ArcFace-Resnet18~\cite{deng_arcface_2019}.
From the results, we can notice that our method achieves the best performance in the LPIPS, FID, and Deg metrics, and the comparable performance in NIQE.
Regarding facial identity preserving performance, our method outperforms the second best method in 9.53\%.
Note that as recent restoration works~\cite{blau20182018, wang_towards_2021} suggested, the pixel-wise metrics (\eg, PSNR and SSIM) are not strongly correlated to visual quality of images, and thus our method is not good at them does not mean our results are worse than others.

In Figure~\ref{fig:realcomp}, we further present the turbulence mitigation results of compared methods in real-world conditions, which are more severely degraded than  the synthesized ones.
However, our method can still achieve its superiority in visual quality comparisons, with fewer unnatural artifacts compared with both the turbulence mitigation methods and BFR methods.
We argue that such superiority comes from the learning objective difference, in which our objective bypasses the uncertainty of datas, and thus it can better preserve the identity features of restored images.

In Table~\ref{tab:very} and Table~\ref{tab:veryall}, we show the pre-trained ArcFace-Resnet18 face recognition accuracy of the restored results, in the center pose, and center, left, right poses, respectively.
From the comparisons, one can easily find that our method significantly outperforms the other methods in recognition accuracy \emph{Top1} and \emph{Top3}, in both the center pose and all poses.
Compared with the second best methods, ours outperform it over 7.86\% and 11.24\%.
Though GFPGAN partly achieved better \emph{Deg} than ours, we find that it tends to produce more unfaithful details that effects \emph{Deg}, but unfaithful details is not identical and lead to a worse \emph{Top1}.

\begin{table}[t]
  \centering
  \resizebox{\linewidth}{!}{
  \begin{tabular}{@{}lcccc@{}}
    \toprule
    & \!Top1 $\uparrow$\! & \!Top3 $\uparrow$\! & \!Top5 $\uparrow$\! & \!Deg. (\%) $\uparrow$\! \\ \midrule
    Turbulence Images & 14.61 & 26.97 & 32.58 & 13.59 \\
    TDRN~\cite{yasarla_learning_2020} [arXiv20] & 13.37 & 28.99 & 30.11 & 10.41 \\
    ATFaceGAN~\cite{lau_atfacegan_2021} [TBIOM21] & 23.60 & 38.20 & 51.69 & 17.26 \\
    PSFRGAN~\cite{chen_progressive_2021} [CVPR21] & \cellcolor{yellow} 39.33 & \cellcolor{yellow} 56.18 & 61.80 & 27.71 \\
    GFPGAN~\cite{wang_towards_2021} [CVPR21] & \cellcolor{orange} 49.44 & \cellcolor{orange} 68.54 & \cellcolor{orange} 79.78 & \cellcolor{yellow} 32.08 \\
    GFPGAN* & 34.83 & 52.81 & \cellcolor{yellow} 64.04 & \cellcolor{orange} 35.11 \\
    LTT-GAN (Ours) & \cellcolor{red} 57.30 & \cellcolor{red} 71.91 & \cellcolor{red} 82.02 & \cellcolor{red} 35.27 \\
    \bottomrule
  \end{tabular}
  }
  \vspace{-0.5\baselineskip}
  \caption{A verification accuracy comparison against the state-of-the-art methods and ours on the real-world turbulence degraded face images collected at 300 meters.}
  \label{tab:very}
  \vspace{-1\baselineskip}
\end{table}

\begin{figure*}[htbp]
  \includegraphics[width=\linewidth]{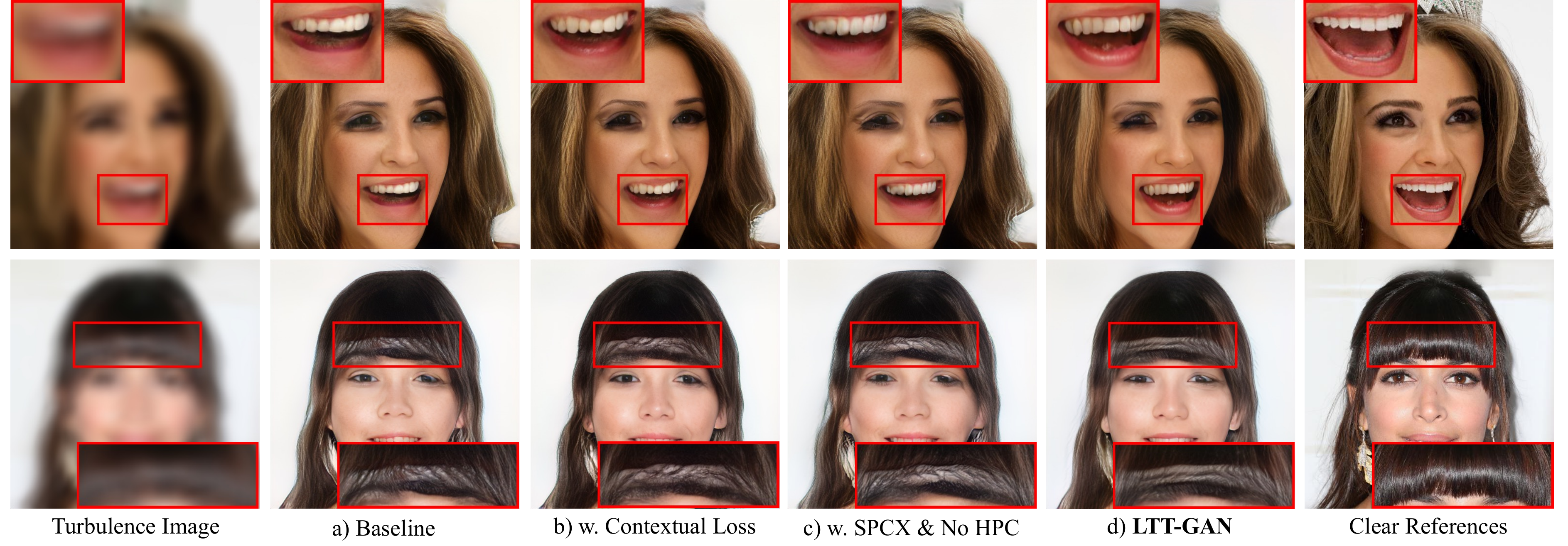}
  \vspace{-1.5\baselineskip}
  \caption{Visualization results of our method on the synthesized CelebAHQ100, as we vary the usage of SPCX and HPC. Compared with the baseline, or the baseline combines with the contextual loss, our method significantly reduce the unnatural artifacts with better details.}
  \label{fig:ablationimg}
  \vspace{-1\baselineskip}
\end{figure*}

\begin{table}[htbp]
  \centering
  \resizebox{\linewidth}{!}{
  \begin{tabular}{@{}lcccc@{}}
    \toprule
    & \!Top1 $\uparrow$\! & \!Top3 $\uparrow$\! & \!Top5 $\uparrow$\! & \!Deg. (\%) $\uparrow$\! \\ \midrule
    Turbulence Images & 14.61 & 44.94 & 50.56 & 11.11 \\
    TDRN~\cite{yasarla_learning_2020} [arXiv20] & 19.10 & 42.11 & 53.68 & 12.46 \\
    ATFaceGAN~\cite{lau_atfacegan_2021} [TBIOM21] & 23.60 & 64.04 & 69.66 & 13.48 \\
    PSFRGAN~\cite{chen_progressive_2021} [CVPR21] & \cellcolor{yellow} 41.57 & 73.03 & 84.27 & 20.30 \\
    GFPGAN~\cite{wang_towards_2021} [CVPR21] & \cellcolor{orange} 48.31 & \cellcolor{orange} 85.39 &  \cellcolor{red} 95.51 & \cellcolor{yellow} 24.57 \\
    GFPGAN* & 37.08 & \cellcolor{yellow} 78.65 & \cellcolor{yellow} 86.52 & \cellcolor{red} 27.46 \\
    LTT-GAN (Ours) & \cellcolor{red} 59.55 & \cellcolor{red} 87.64 & \cellcolor{orange} 93.26 & \cellcolor{orange} 26.57 \\
    \bottomrule
  \end{tabular}
  }
  \vspace{-0.5\baselineskip}
  \caption{A verification accuracy comparison against the state-of-the-art methods and ours on the real-world turbulence degraded face images collected at 300 meters in center, left, and right poses.}
  \label{tab:veryall}
  \vspace{-1\baselineskip}
\end{table}

\begin{table}[htbp]
  \centering
  \resizebox{\linewidth}{!}{
  \begin{tabular}{@{}lccc@{}}
    \toprule
    & \!LPIPS $\downarrow$\! & \!FID $\downarrow$\! & \!Deg. (\%) $\uparrow$\! \\
    \midrule
    no./\emph{StyleG} no./\emph{StyleD} no./\emph{ID}  & 0.3765 & 134.15 & 23.64 \\
    no./\emph{StyleG} no./\emph{StyleD}  & 0.3733 & 114.56 & 33.02 \\
    no./\emph{StyleG} & 0.3720 & 114.23 & 33.19 \\
    \hline
    Baseline & 0.3469 & 85.36 & 36.80 \\
    \hline
    w./\emph{Brute-Force $\mathcal{L}_2$} & \cellcolor{orange} 0.3357 & 86.62 & 36.70 \\
    w./\emph{Brute-Force ID} & 0.3510 & 95.72 & 37.77 \\
    w./\emph{CX} & \cellcolor{yellow} 0.3358 & \cellcolor{orange} 85.08 & 38.57 \\
    w./\emph{SPCX(r=8)} & 0.3360 & 87.94 & 38.10 \\
    w./\emph{SPCX(r=16)} & \cellcolor{red} 0.3328 & \cellcolor{red} 82.75 & \cellcolor{yellow} 38.67 \\
    w./\emph{SPCX(r=32)} & 0.3370 & \cellcolor{yellow} 85.90 & \cellcolor{red} 39.20 \\
    w./\emph{SPCX(r=64)} & 0.3403 & 91.54 & \cellcolor{orange} 38.90 \\
    \hline
    w./\emph{$\mathcal{L}_2$} w./\emph{HPC(g=3)} & 0.3429 & \cellcolor{red} 87.43 & 38.75 \\
    w./\emph{SPCX(r=32)} w./\emph{HPC(g=3)} & \cellcolor{red} 0.3273 & \cellcolor{orange} 92.80 & \cellcolor{yellow} 40.03 \\
    w./\emph{SPCX(r=32)} w./\emph{HPC(g=4)} &  \cellcolor{orange} 0.3324 & \cellcolor{yellow} 94.22 & \cellcolor{red} 40.74 \\
    w./\emph{SPCX(r=32)} w./\emph{HPC(g=5)} & \cellcolor{yellow} 0.3395 & 102.64 & \cellcolor{orange} 40.57 \\
    \bottomrule
  \end{tabular}
  }
  \vspace{-0.5\baselineskip}
  \caption{A performance comparison against several different settings of our method on the synthesized CelebAHQ100.}
  \label{tab:abl}
  \vspace{-1\baselineskip}
\end{table}

\subsection{Ablations and Discussions}
In this section, we conduct experiments on different settings of our method to discuss their effectiveness on the turbulence mitigation capability, includes \emph{Baseline ($\mathcal{L}_2$ for identity)} \vs \emph{Baseline with Brute-Force $\mathcal{L}_2$ (larger weights of $\mathcal{L}_2$)} \vs \emph{Baseline with Brute-Force ID (larger weights of $ID$)} \vs \emph{Baseline with Contextual loss~\cite {mechrez_contextual_2018}} (CX) \vs \emph{LTT-GAN with different rates of SPCX} \vs \emph{LTT-GAN with SPCX(r=32) and different group sizes of HPC}.
To simplify, we train each method with 80k iterations only, and thus its numbers differ from what is reported in Table~\ref{tab:syncomp} in 800k iterations.

The comparisons are shown in Table~\ref{tab:abl} and Figure~\ref{fig:ablationimg}.
Compared with the baseline and several variants of our method, our proposed SPCX and HCP can consistently reduce unnatural artifacts in images, and the final results from our LTT-GAN significantly achieve the best identity compared with the clean reference images.
Since the rate $r$ of SPCX depends on the minimal dimension of identity features that can denote a single face, the optimal value may varies depending on the datasets.
The application of HPC also significantly increases the performance in all metrics.
Please refer to the supplement for the limitation discussion.

\section{Conclusion}
We have presented LTT-GAN, a new generative embedding network that can achieve the best trade-off between the identity and realism of restored turbulence images.
Compared to previous approaches, LTT-GAN achieves the goal in a similar way of the adversarial loss, but instead, is implemented with a new contextual distance.
The new objective can better consider the identity difference than previously applied losses.
To further boost its performance, we modify the embedded well-trained GAN to allow multiple results to be acquired from a single input, which can introduce more appearance variance without identity changing in a single forward pass.
On the difficult turbulence mitigation problem, LTT-GAN is the first GAN inversion method, and it provided significant performance improvements over the existing state-of-the-art methods in both the synthesized and real-world turbulence degraded images.

% %%%%%%%%% REFERENCES
{\small
\bibliographystyle{ieee_fullname}
\bibliography{GANInversion}
}

\newpage
\appendix
\section{Limitations}
\noindent \textbf{Training Data Bias.}
The applied training dataset follows the most common settings of the face generation tasks, \ie, FFHQ.
However, we find that the dataset has data bias towards dark-skinned faces.
The network learns with the bias tends to light the face images when the degradation is severe, \eg, real-world turbulence degradation.
In the following figure, we show our turbulence mitigation results in both the synthesized and real-world turbulence images, respectively.
From the results, we suggest that the way of solving the issue is including more diverse images at training, or applying a controllable synthetic dataset.

\begin{figure}[h]
    \begin{subfigure}{\linewidth}
        \includegraphics[width=.24\linewidth]{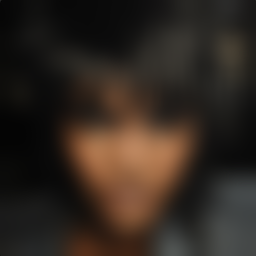}
        \includegraphics[width=.24\linewidth]{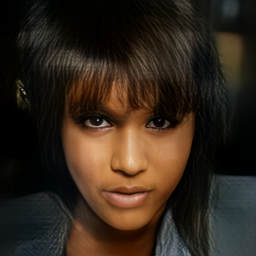}
        \includegraphics[width=.24\linewidth]{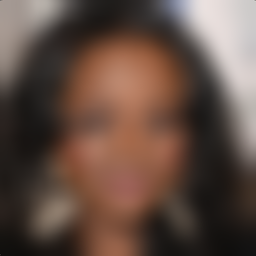}
        \includegraphics[width=.24\linewidth]{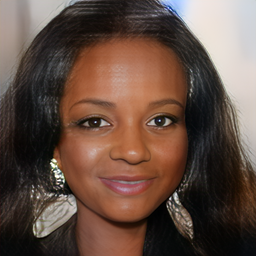}
        \caption{Synthesized turbulence images with dark-skinned faces and our results.}
    \end{subfigure}
     \begin{subfigure}{\linewidth}
        \includegraphics[width=.24\linewidth]{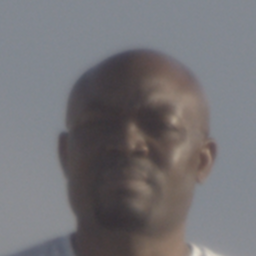}
        \includegraphics[width=.24\linewidth]{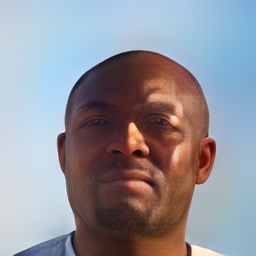}
        \includegraphics[width=.24\linewidth]{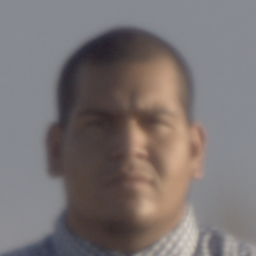}
        \includegraphics[width=.24\linewidth]{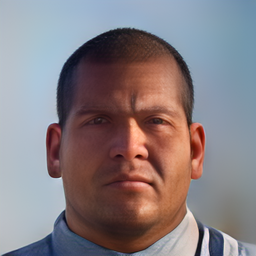}
        \caption{Real-world turbulence images with dark-skinned faces and our results.}
    \end{subfigure}
\end{figure}

\noindent \textbf{Spatial Inductive Bias.}
As recent works suggested, most state-of-the-art generative networks, \eg, StyleGAN, have strong spatial inductive bias.
Specifically, the fine details appear to be fixed in pixel coordinates.
We empirically find that such a spatial inductive bias also affects the restored results of GAN inversion.
In the following figure, we show our turbulence mitigation results as well as the baseline, \ie, GFPGAN in three different poses.
Thanks to the proposed spatial periodic contextual distance, which can bypass the spatial pixel difference, our results are significantly more robust than GFPGAN with less unnatural artifacts.
Though these unnatural artifacts led by the bias have limited effects on face recognition accuracy.
However, our results in left and right poses are less realistic than the results in center poses.

\begin{figure}[h]
    \begin{subfigure}{\linewidth}
        \includegraphics[width=.24\linewidth]{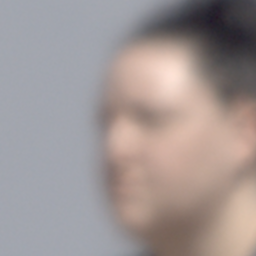}
        \includegraphics[width=.24\linewidth]{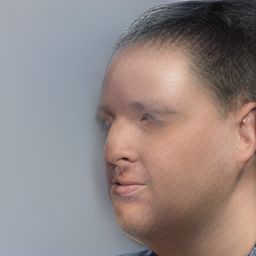}
        \includegraphics[width=.24\linewidth]{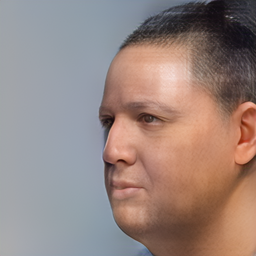}
        \includegraphics[width=.24\linewidth]{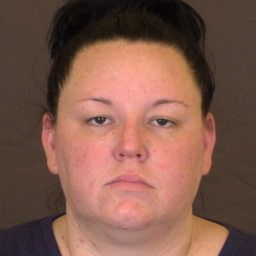}
    \end{subfigure}
    \begin{subfigure}{\linewidth}
        \includegraphics[width=.24\linewidth]{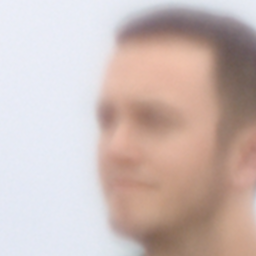}
        \includegraphics[width=.24\linewidth]{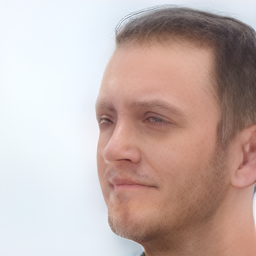}
        \includegraphics[width=.24\linewidth]{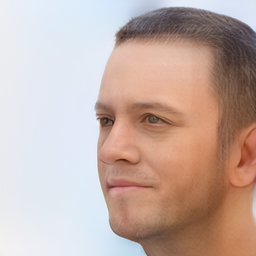}
        \includegraphics[width=.24\linewidth]{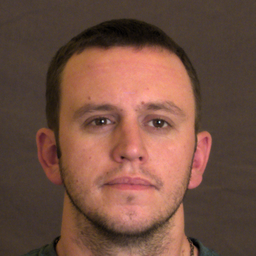}
        \caption{Turbulence Images, GFPGAN results, our results, references.}
    \end{subfigure}
\end{figure}

\section{Uncertainty Visualization}
\begin{figure}[h]
    \centering
    \begin{subfigure}[c]{.23\linewidth}
      \captionsetup{justification=centering, labelformat=empty, font=scriptsize}
      \includegraphics[width=\linewidth]{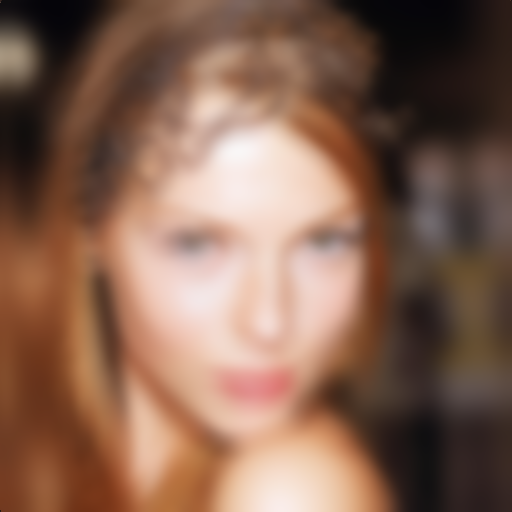}
      \caption{Turbulence Image}
    \end{subfigure}
    \begin{subfigure}[c]{.23\linewidth}
      \captionsetup{justification=centering, labelformat=empty, font=scriptsize}
      \includegraphics[width=\linewidth]{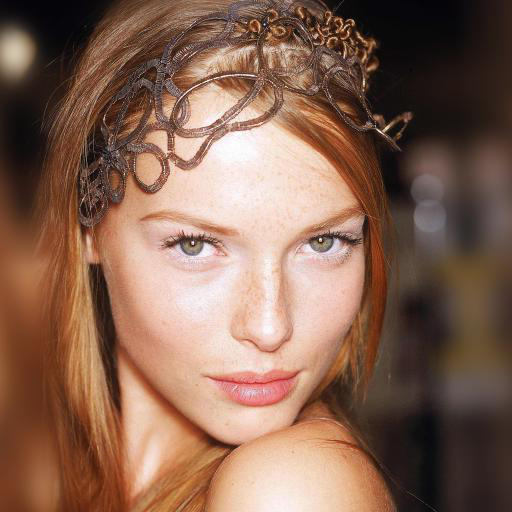}
      \caption{Reference}
    \end{subfigure}
    \begin{subfigure}[c]{.23\linewidth}
      \captionsetup{justification=centering, labelformat=empty, font=scriptsize}
      \includegraphics[width=\linewidth]{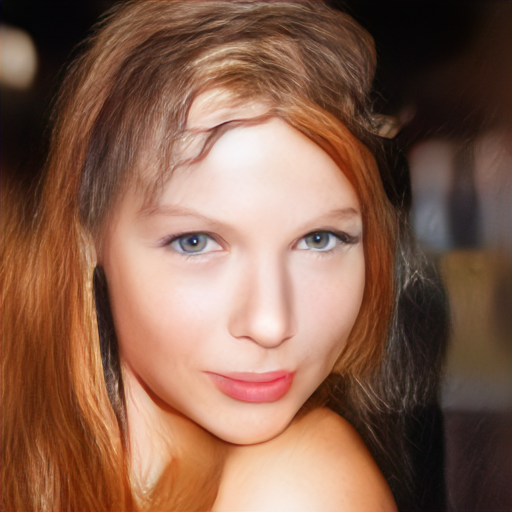}
      \caption{LTT-GAN Result}
    \end{subfigure}
    \begin{subfigure}[c]{.26\linewidth}
      \captionsetup{justification=centering, labelformat=empty, font=scriptsize}
      \includegraphics[width=\linewidth]{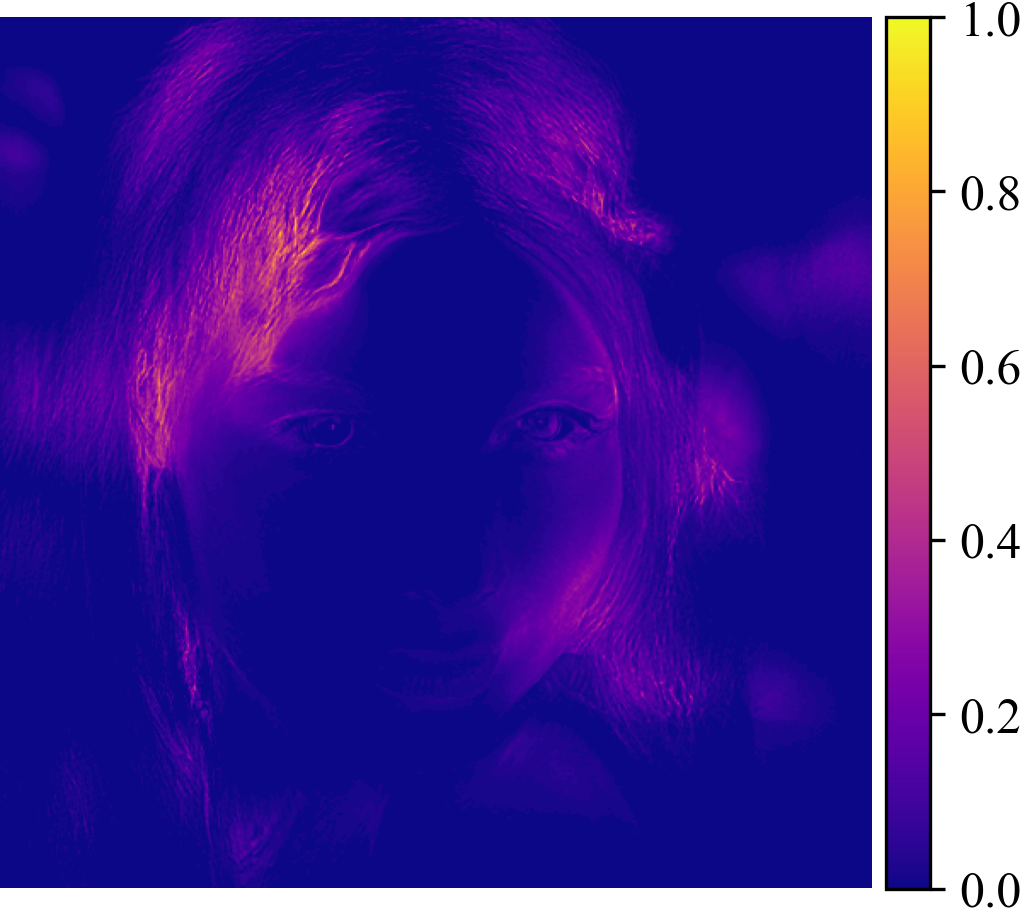}
      \caption{Uncertanity Map}
    \end{subfigure}
    \caption{Visualization of the result variance map. Even the network generates a sharp result with hair patterns in the original hair band area, the uncertainty map can identify the area is uncertain.}
    \label{fig:uncer}
\end{figure}

Identifying the uncertainty of restored results can help both the manual inspection and automated methods.
Conventional restoration methods tend to produce artifacts in uncertain areas, which can help to identify.
However, in GAN inversion, the uncertainty area of their results remains sharp, and hence it is hard to identify, but the imperceptible uncertainty is extremely dangerous in practical applications.
With the help of HPC and SPCX, our method now can show the uncertainty map of its prediction using the variance of its pseudo results, as shown in Figure~\ref{fig:uncer}.
By simply measuring the variance of multiple pseudo result images, we can treat the uncertainty as the variance of each pixel, in a single forward pass.
It does not require additional network architecture modification, \eg, Monte Carlo dropouts~\cite{kendall2017uncertainties} used in TDRN~\cite{yasarla_learning_2020}, nor multiple forward passes of Style-Mixing used in recent GAN inversion methods~\cite{shen2021closed, richardson_encoding_2021}.

\section{Code}
Algorithm~\ref{alg:code} provides reference NumPy code for $\mathcal{P}\mathcal{K}$ operation and spatial periodic contextual distance measurement.
A comparable code for vanilla contextual distance is provided for reference too.
A complete code release will be made available upon publication.

\section{Additional Results}
\begin{figure*}[htbp]
    \begin{subfigure}[t]{.12\linewidth}
      \captionsetup{justification=centering, labelformat=empty, font=scriptsize}
      \includegraphics[width=1\linewidth]{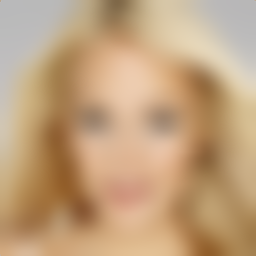}
      \includegraphics[width=1\linewidth]{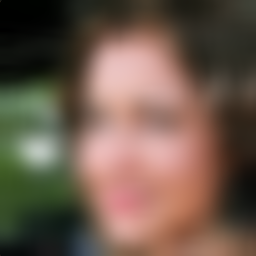}
      \includegraphics[width=1\linewidth]{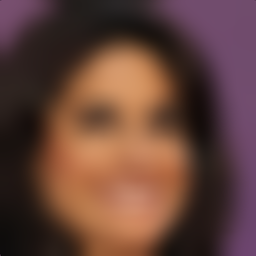}
      \includegraphics[width=1\linewidth]{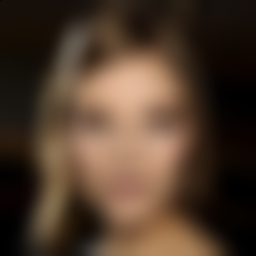}
      \includegraphics[width=1\linewidth]{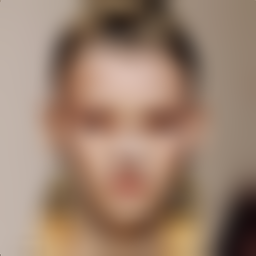}
      \includegraphics[width=1\linewidth]{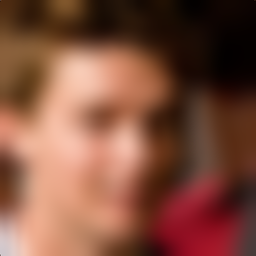}
      \includegraphics[width=1\linewidth]{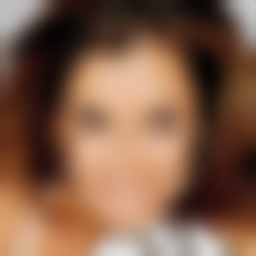}
      \includegraphics[width=1\linewidth]{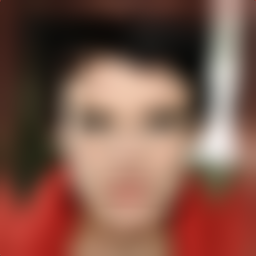}
      \includegraphics[width=1\linewidth]{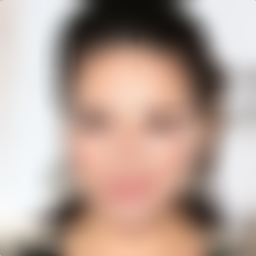}
      \caption{Turbulence Image}
    \end{subfigure}
    \begin{subfigure}[t]{.12\linewidth}
      \captionsetup{justification=centering, labelformat=empty, font=scriptsize}
      \includegraphics[width=1\linewidth]{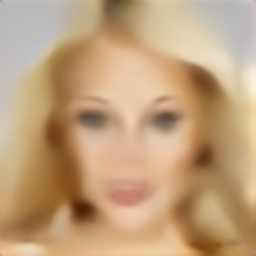}
      \includegraphics[width=1\linewidth]{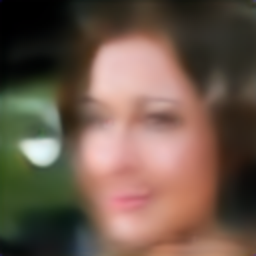}
      \includegraphics[width=1\linewidth]{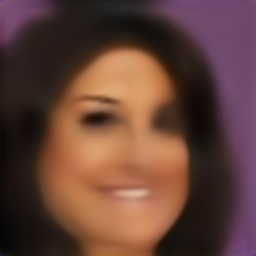}
      \includegraphics[width=1\linewidth]{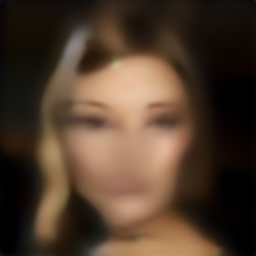}
      \includegraphics[width=1\linewidth]{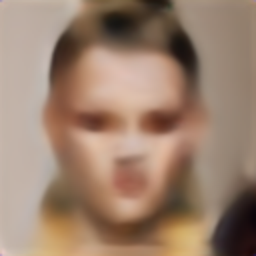}
      \includegraphics[width=1\linewidth]{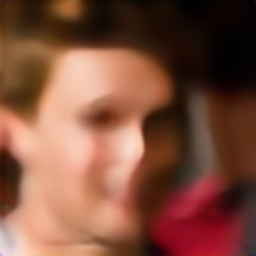}
      \includegraphics[width=1\linewidth]{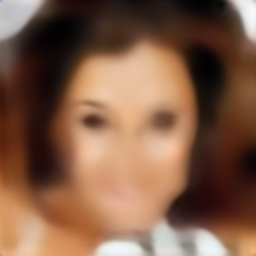}
      \includegraphics[width=1\linewidth]{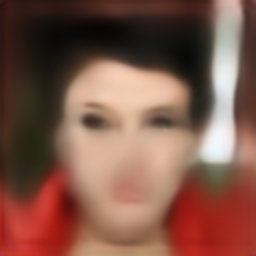}
      \includegraphics[width=1\linewidth]{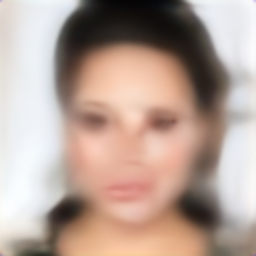}
      \caption{TDRN \\ \emph{arXiv20}}
    \end{subfigure}
    \begin{subfigure}[t]{.12\linewidth}
      \captionsetup{justification=centering, labelformat=empty, font=scriptsize}
      \includegraphics[width=1\linewidth]{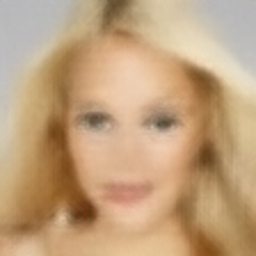}
      \includegraphics[width=1\linewidth]{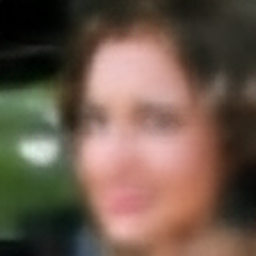}
      \includegraphics[width=1\linewidth]{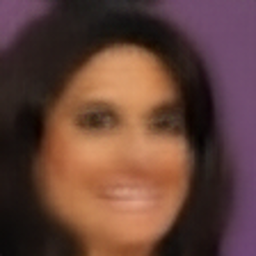}
      \includegraphics[width=1\linewidth]{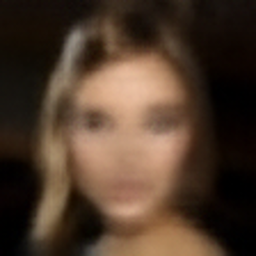}
      \includegraphics[width=1\linewidth]{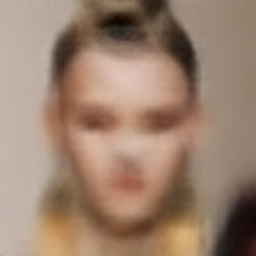}
      \includegraphics[width=1\linewidth]{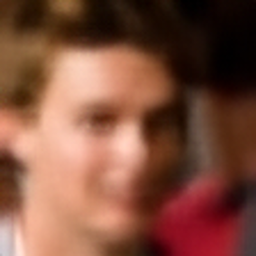}
      \includegraphics[width=1\linewidth]{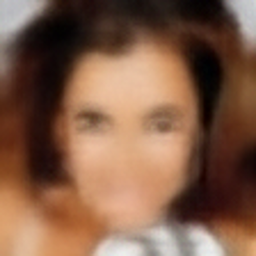}
      \includegraphics[width=1\linewidth]{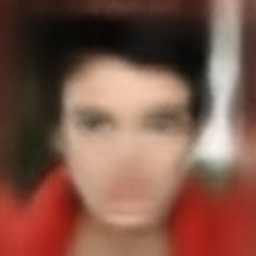}
      \includegraphics[width=1\linewidth]{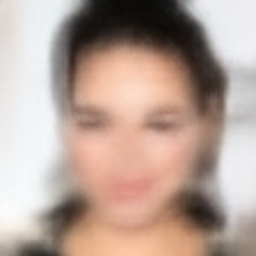}
      \caption{ATFaceGAN \\ \emph{TBIOM21}}
    \end{subfigure}
    \begin{subfigure}[t]{.12\linewidth}
      \captionsetup{justification=centering, labelformat=empty, font=scriptsize}
      \includegraphics[width=1\linewidth]{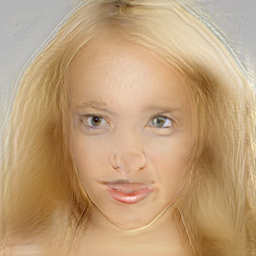}
      \includegraphics[width=1\linewidth]{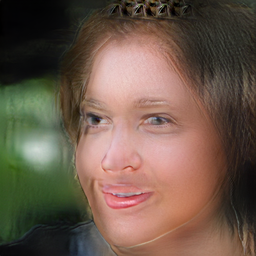}
      \includegraphics[width=1\linewidth]{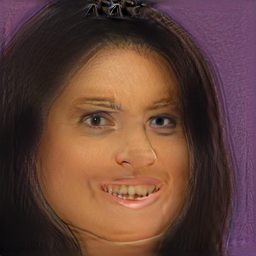}
      \includegraphics[width=1\linewidth]{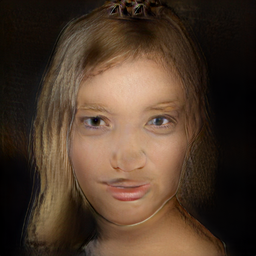}
      \includegraphics[width=1\linewidth]{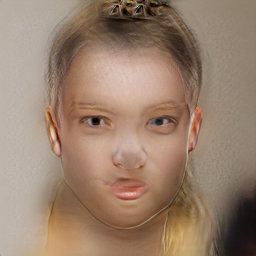}
      \includegraphics[width=1\linewidth]{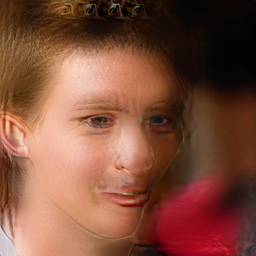}
      \includegraphics[width=1\linewidth]{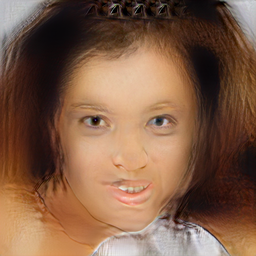}
      \includegraphics[width=1\linewidth]{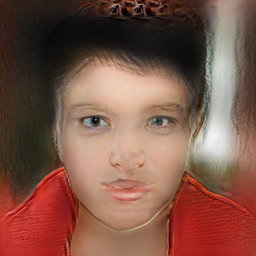}
      \includegraphics[width=1\linewidth]{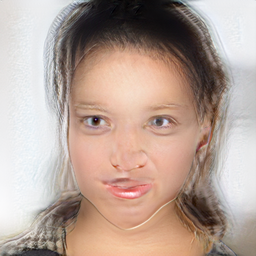}
      \caption{PSFRGAN \\ \emph{CVPR21}}
    \end{subfigure}
    \begin{subfigure}[t]{.12\linewidth}
      \captionsetup{justification=centering, labelformat=empty, font=scriptsize}
      \includegraphics[width=1\linewidth]{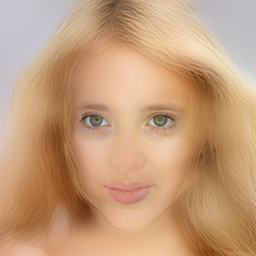}
      \includegraphics[width=1\linewidth]{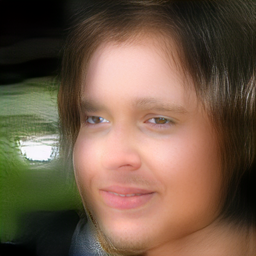}
      \includegraphics[width=1\linewidth]{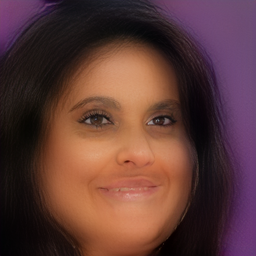}
      \includegraphics[width=1\linewidth]{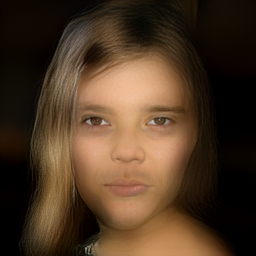}
      \includegraphics[width=1\linewidth]{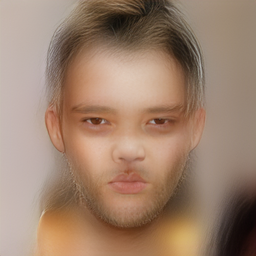}
      \includegraphics[width=1\linewidth]{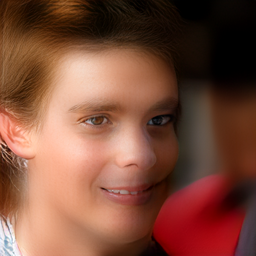}
      \includegraphics[width=1\linewidth]{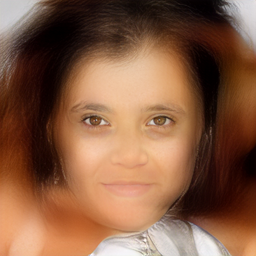}
      \includegraphics[width=1\linewidth]{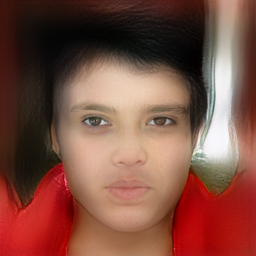}
      \includegraphics[width=1\linewidth]{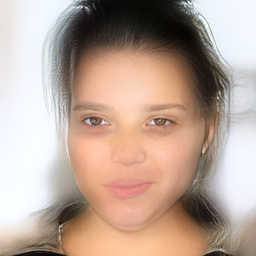}
      \caption{GFPGAN \\ \emph{CVPR21}}
    \end{subfigure}
    \begin{subfigure}[t]{.12\linewidth}
      \captionsetup{justification=centering, labelformat=empty, font=scriptsize}
      \includegraphics[width=1\linewidth]{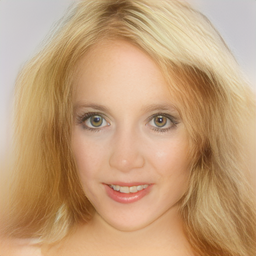}
      \includegraphics[width=1\linewidth]{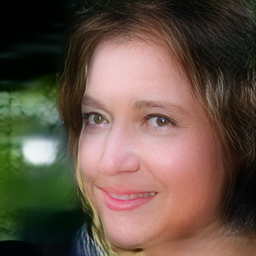}
      \includegraphics[width=1\linewidth]{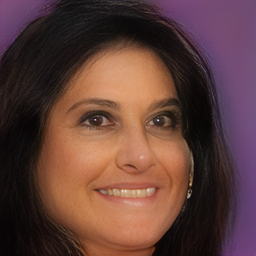}
      \includegraphics[width=1\linewidth]{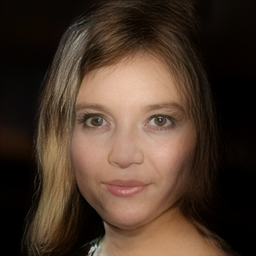}
      \includegraphics[width=1\linewidth]{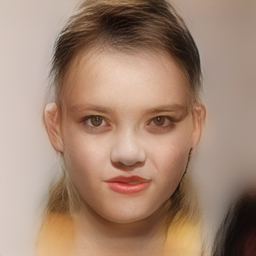}
      \includegraphics[width=1\linewidth]{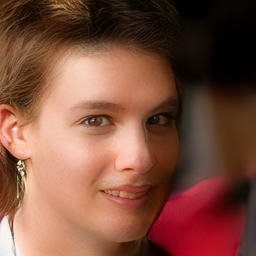}
      \includegraphics[width=1\linewidth]{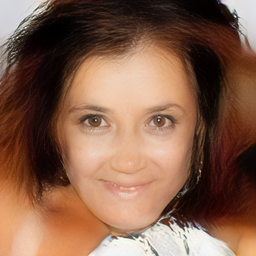}
      \includegraphics[width=1\linewidth]{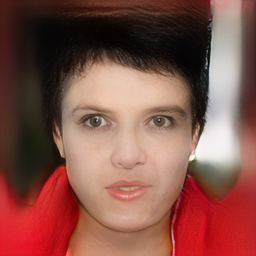}
      \includegraphics[width=1\linewidth]{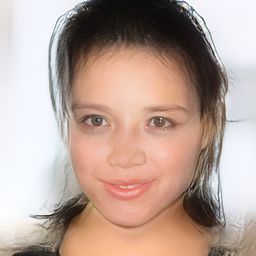}
      \caption{GFPGAN* \\ (ElasticAug)}
    \end{subfigure}
    \begin{subfigure}[t]{.12\linewidth}
      \captionsetup{justification=centering, labelformat=empty, font=scriptsize}
      \includegraphics[width=1\linewidth]{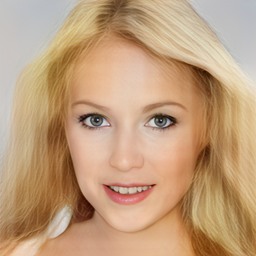}
      \includegraphics[width=1\linewidth]{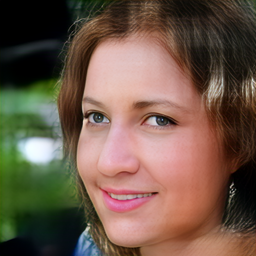}
      \includegraphics[width=1\linewidth]{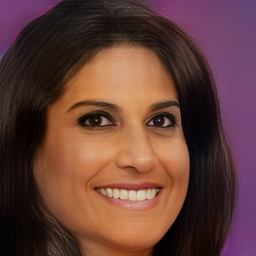}
      \includegraphics[width=1\linewidth]{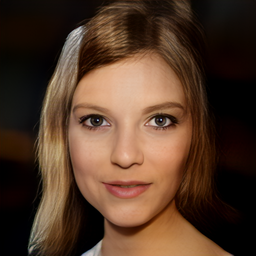}
      \includegraphics[width=1\linewidth]{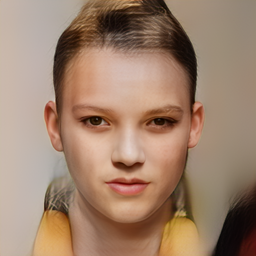}
      \includegraphics[width=1\linewidth]{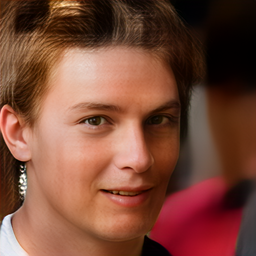}
      \includegraphics[width=1\linewidth]{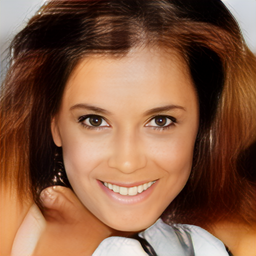}
      \includegraphics[width=1\linewidth]{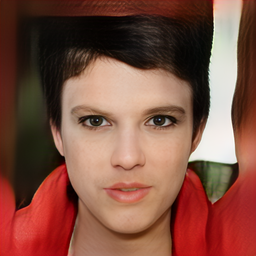}
      \includegraphics[width=1\linewidth]{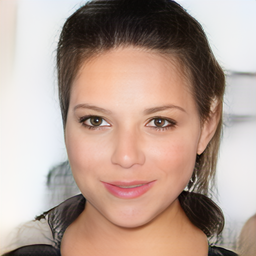}
      \caption{\textbf{LTT-GAN \\ \emph{Ours}}}
    \end{subfigure}
    \begin{subfigure}[t]{.12\linewidth}
      \captionsetup{justification=centering, labelformat=empty, font=scriptsize}
      \includegraphics[width=1\linewidth]{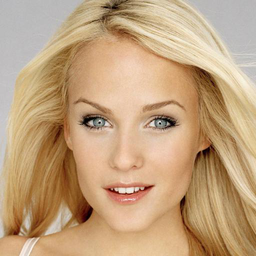}
      \includegraphics[width=1\linewidth]{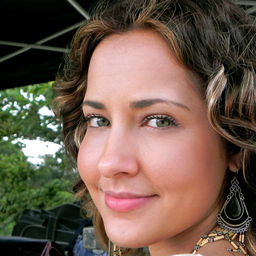}
      \includegraphics[width=1\linewidth]{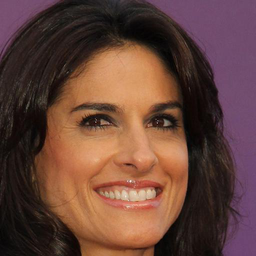}
      \includegraphics[width=1\linewidth]{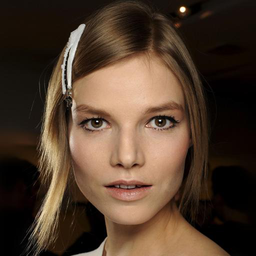}
      \includegraphics[width=1\linewidth]{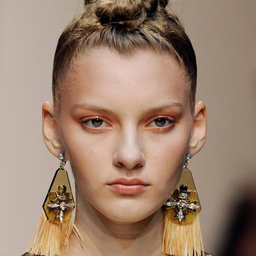}
      \includegraphics[width=1\linewidth]{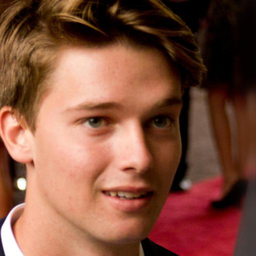}
      \includegraphics[width=1\linewidth]{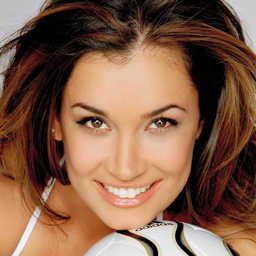}
      \includegraphics[width=1\linewidth]{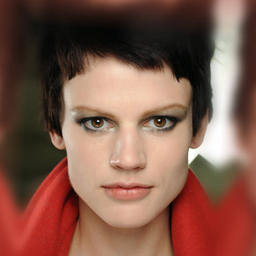}
      \includegraphics[width=1\linewidth]{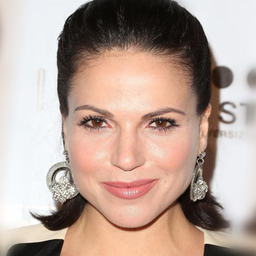}
      \caption{Ground Truth \\ Image}
    \end{subfigure}
    \hfill
    \vspace{-0.5\baselineskip}
    \caption{Additional visualizations of restored synthetic turbulence face images. Compared with results contain unnature artifacts, our results achieve the best visual quality and identity similarity of the ground truth. \textbf{(200\% Zoom is recommended to see their difference)}}
    \label{fig:supsyncomp}
    \vspace{-1\baselineskip}
  \end{figure*}

\begin{figure*}
    \begin{subfigure}[t]{.12\linewidth}
      \captionsetup{justification=centering, labelformat=empty, font=scriptsize}
      \includegraphics[width=1\linewidth]{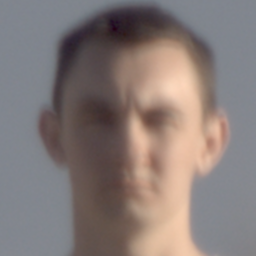}
      \includegraphics[width=1\linewidth]{figures/supplement/real/tubimages/010.png}
      \includegraphics[width=1\linewidth]{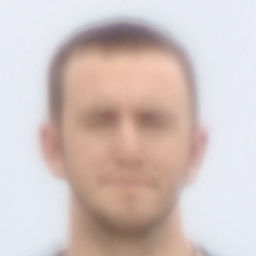}
      \includegraphics[width=1\linewidth]{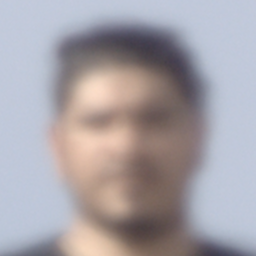}
      \includegraphics[width=1\linewidth]{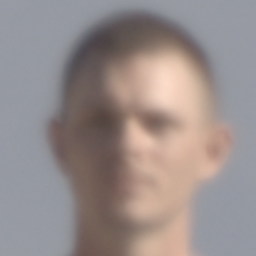}
      \includegraphics[width=1\linewidth]{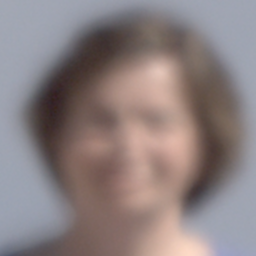}
      \includegraphics[width=1\linewidth]{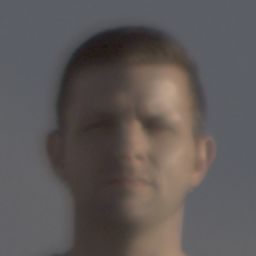}
      \includegraphics[width=1\linewidth]{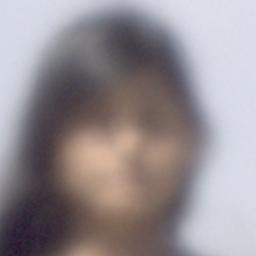}
      \includegraphics[width=1\linewidth]{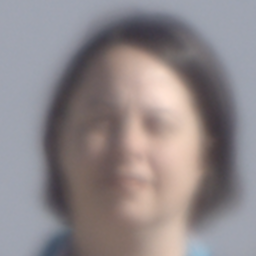}
      \caption{Turbulence Image}
    \end{subfigure}
    \begin{subfigure}[t]{.12\linewidth}
      \captionsetup{justification=centering, labelformat=empty, font=scriptsize}
      \includegraphics[width=1\linewidth]{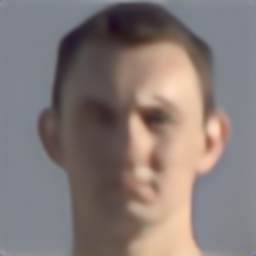}
      \includegraphics[width=1\linewidth]{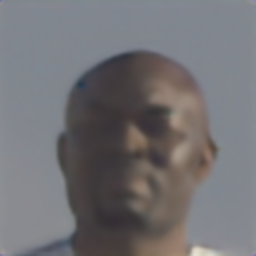}
      \includegraphics[width=1\linewidth]{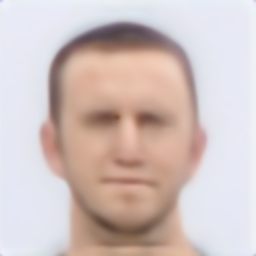}
      \includegraphics[width=1\linewidth]{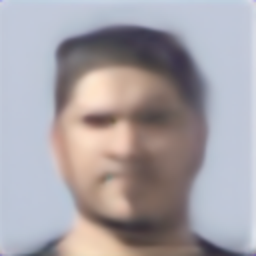}
      \includegraphics[width=1\linewidth]{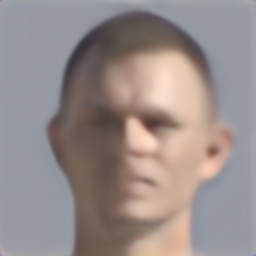}
      \includegraphics[width=1\linewidth]{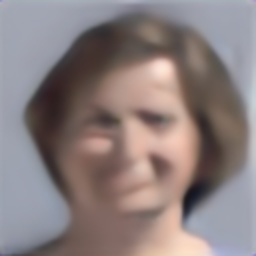}
      \includegraphics[width=1\linewidth]{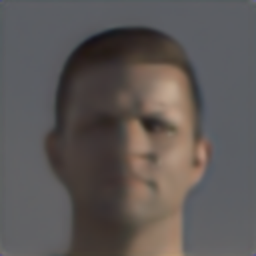}
      \includegraphics[width=1\linewidth]{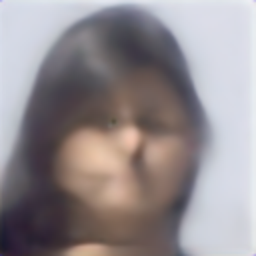}
      \includegraphics[width=1\linewidth]{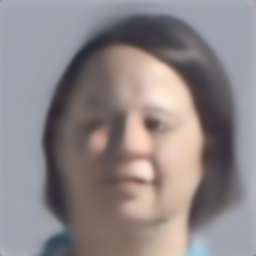}
      \caption{TDRN \\ \emph{arXiv20}}
    \end{subfigure}
    \begin{subfigure}[t]{.12\linewidth}
      \captionsetup{justification=centering, labelformat=empty, font=scriptsize}
      \includegraphics[width=1\linewidth]{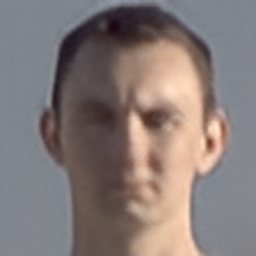}
      \includegraphics[width=1\linewidth]{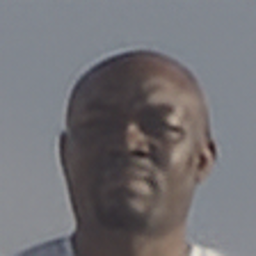}
      \includegraphics[width=1\linewidth]{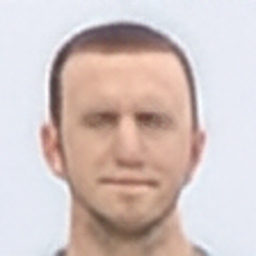}
      \includegraphics[width=1\linewidth]{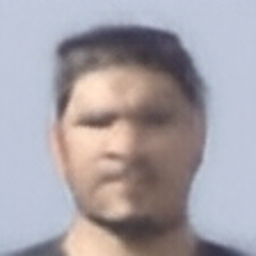}
      \includegraphics[width=1\linewidth]{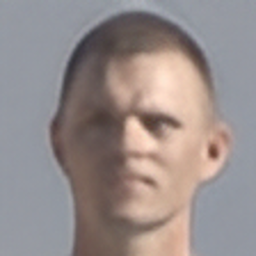}
      \includegraphics[width=1\linewidth]{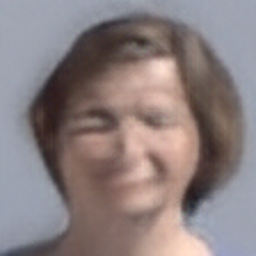}
      \includegraphics[width=1\linewidth]{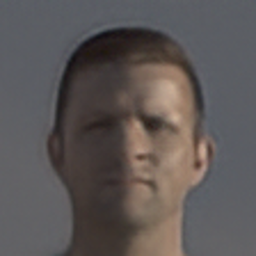}
      \includegraphics[width=1\linewidth]{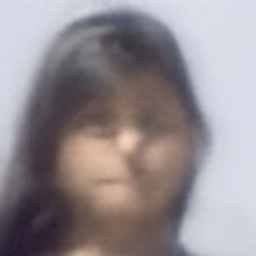}
      \includegraphics[width=1\linewidth]{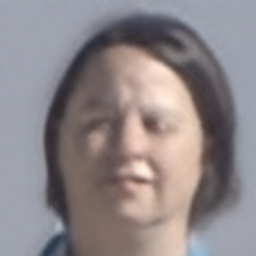}
      \caption{ATFaceGAN \\ \emph{TBIOM21}}
    \end{subfigure}
    \begin{subfigure}[t]{.12\linewidth}
      \captionsetup{justification=centering, labelformat=empty, font=scriptsize}
      \includegraphics[width=1\linewidth]{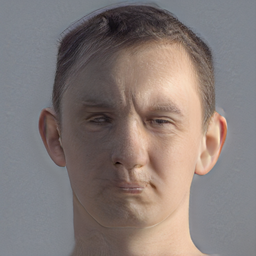}
      \includegraphics[width=1\linewidth]{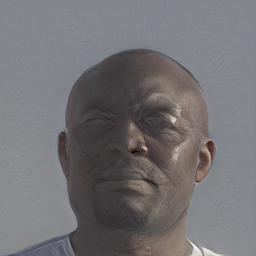}
      \includegraphics[width=1\linewidth]{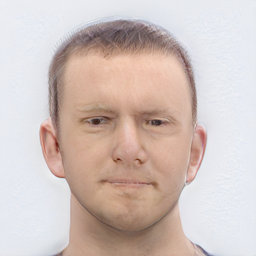}
      \includegraphics[width=1\linewidth]{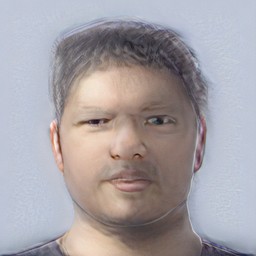}
      \includegraphics[width=1\linewidth]{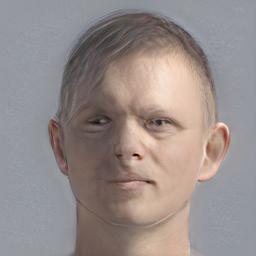}
      \includegraphics[width=1\linewidth]{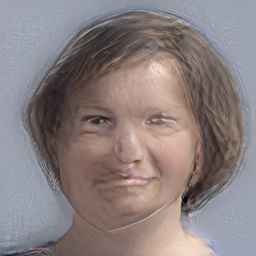}
      \includegraphics[width=1\linewidth]{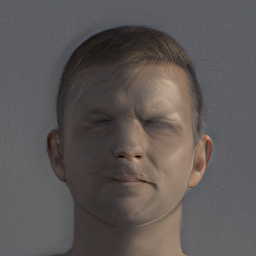}
      \includegraphics[width=1\linewidth]{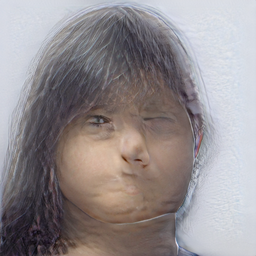}
      \includegraphics[width=1\linewidth]{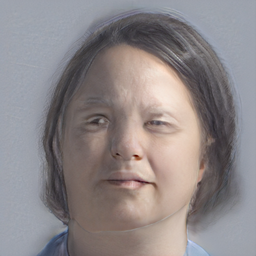}
      \caption{PSFRGAN \\ \emph{CVPR21}}
    \end{subfigure}
    \begin{subfigure}[t]{.12\linewidth}
      \captionsetup{justification=centering, labelformat=empty, font=scriptsize}
      \includegraphics[width=1\linewidth]{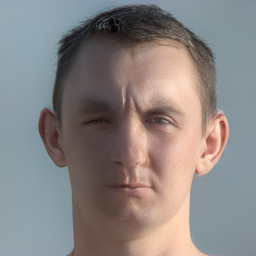}
      \includegraphics[width=1\linewidth]{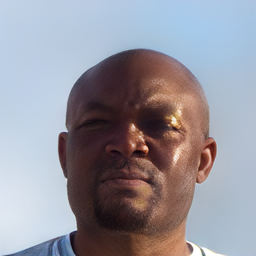}
      \includegraphics[width=1\linewidth]{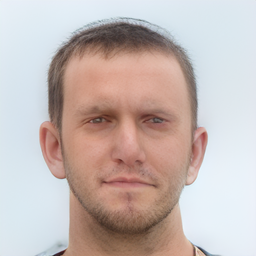}
      \includegraphics[width=1\linewidth]{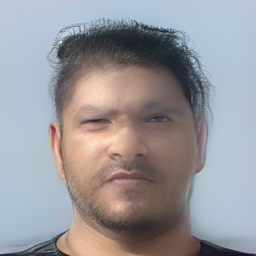}
      \includegraphics[width=1\linewidth]{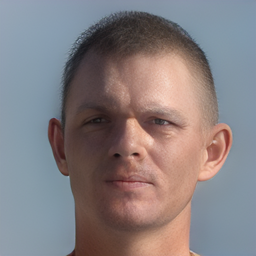}
      \includegraphics[width=1\linewidth]{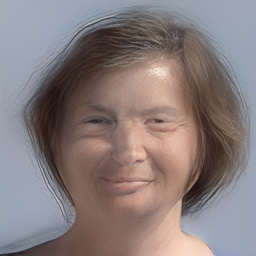}
      \includegraphics[width=1\linewidth]{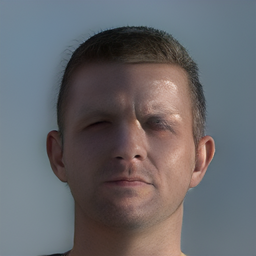}
      \includegraphics[width=1\linewidth]{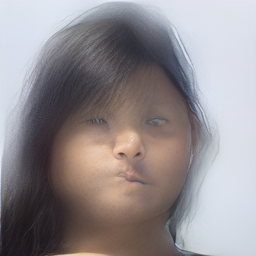}
      \includegraphics[width=1\linewidth]{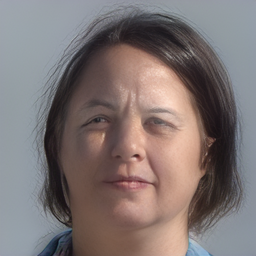}
      \caption{GFPGAN \\ \emph{CVPR21}}
    \end{subfigure}
    \begin{subfigure}[t]{.12\linewidth}
      \captionsetup{justification=centering, labelformat=empty, font=scriptsize}
      \includegraphics[width=1\linewidth]{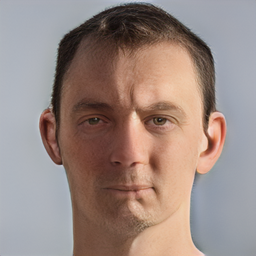}
      \includegraphics[width=1\linewidth]{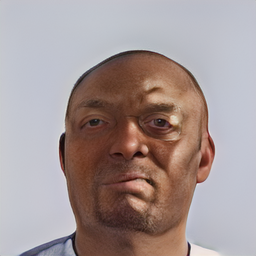}
      \includegraphics[width=1\linewidth]{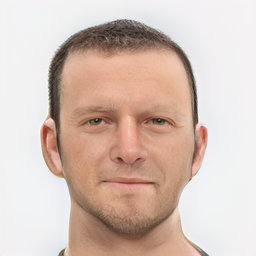}
      \includegraphics[width=1\linewidth]{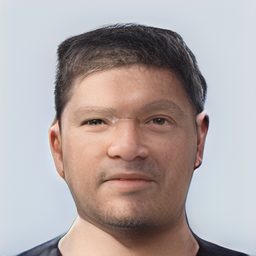}
      \includegraphics[width=1\linewidth]{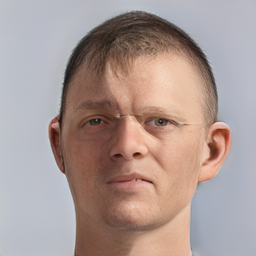}
      \includegraphics[width=1\linewidth]{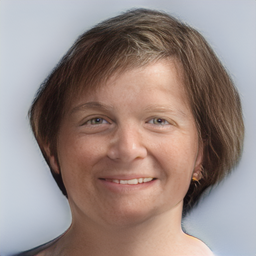}
      \includegraphics[width=1\linewidth]{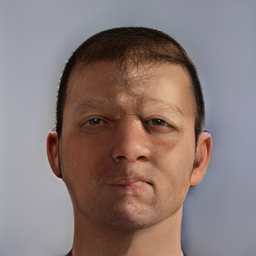}
      \includegraphics[width=1\linewidth]{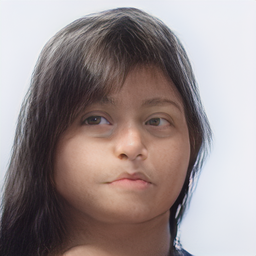}
      \includegraphics[width=1\linewidth]{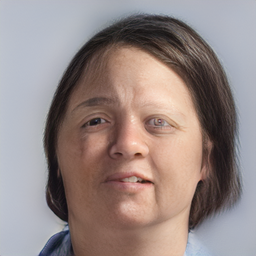}
      \caption{GFPGAN* \\ (ElasticAug)}
    \end{subfigure}
    \begin{subfigure}[t]{.12\linewidth}
      \captionsetup{justification=centering, labelformat=empty, font=scriptsize}
      \includegraphics[width=1\linewidth]{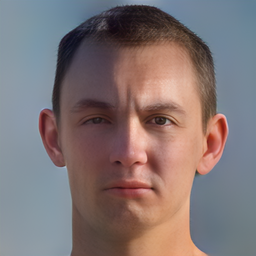}
      \includegraphics[width=1\linewidth]{figures/supplement/real/comparisons/EiGEN_145K/010.png}
      \includegraphics[width=1\linewidth]{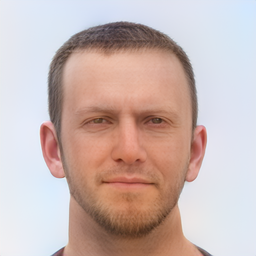}
      \includegraphics[width=1\linewidth]{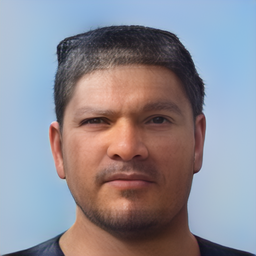}
      \includegraphics[width=1\linewidth]{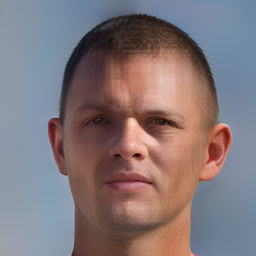}
      \includegraphics[width=1\linewidth]{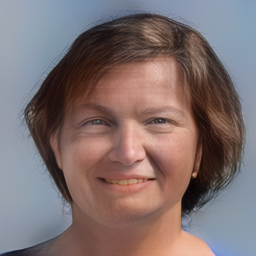}
      \includegraphics[width=1\linewidth]{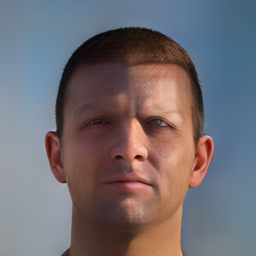}
      \includegraphics[width=1\linewidth]{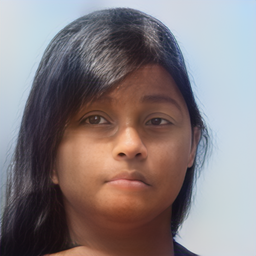}
      \includegraphics[width=1\linewidth]{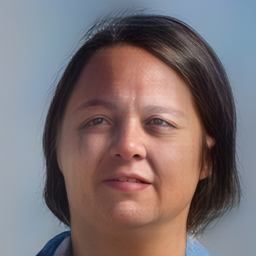}
      \caption{\textbf{LTT-GAN \\ \emph{Ours}}}
    \end{subfigure}
    \begin{subfigure}[t]{.12\linewidth}
      \captionsetup{justification=centering, labelformat=empty, font=scriptsize}
      \includegraphics[width=1\linewidth]{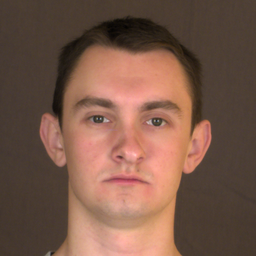}
      \includegraphics[width=1\linewidth]{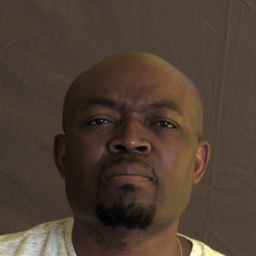}
      \includegraphics[width=1\linewidth]{figures/supplement/real/references/07.png}
      \includegraphics[width=1\linewidth]{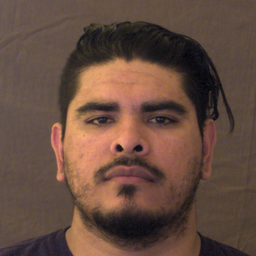}
      \includegraphics[width=1\linewidth]{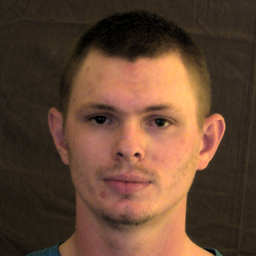}
      \includegraphics[width=1\linewidth]{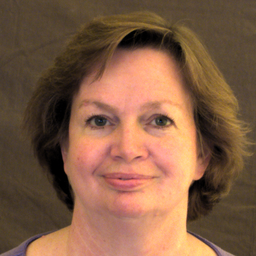}
      \includegraphics[width=1\linewidth]{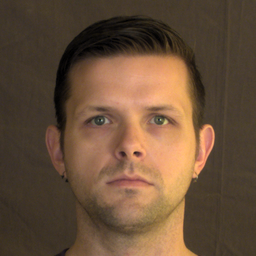}
      \includegraphics[width=1\linewidth]{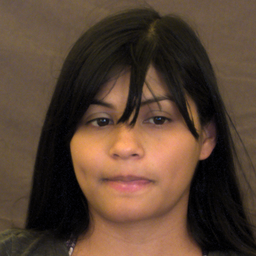}
      \includegraphics[width=1\linewidth]{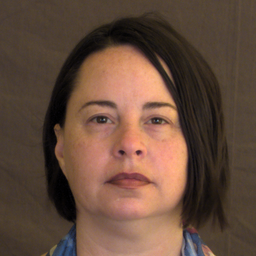}
      \caption{Reference Image}
    \end{subfigure}
    \hfill
    \vspace{-0.5\baselineskip}
    \caption{Additional visualizations of restored real-world turbulence face images. Compared with results contain unnature artifacts, our results achieve the best visual quality and identity similarity of the reference images. \textbf{(200\% Zoom is recommended to see their difference)}}
    \label{fig:suprealcomp}
    \vspace{-1\baselineskip}
  \end{figure*}

Figure~\ref{fig:supsyncomp} presents the additional results of compared methods and ours on the synthetic CelebAHQ100 dataset.
Figure~\ref{fig:suprealcomp} presents the additional results of compared method and ours on the real-world turbulence face TubFace89 dataset.

  \lstset{basicstyle=\ttfamily\small}
  \renewcommand{\tablename}{Algorithm}
  \setcounter{table}{0}
  \begin{table*}[b] 
      \begin{lstlisting}
      import numpy as np
      from VGG import vgg19_feature_extractor
      
      def pk_operation(img, r):
          H, W, C = img.shape
          img = img.reshape(H//r, r, W//r, r, C)
          img = img.transpose(0, 2, 1, 3, 4)
          img = img.reshape(H//r * W//r, r * r * C)
          return img
      
      # vanilla contextual distance
      def CX(X, Y, band_width=0.2):
          perceptual_x = vgg19_feature_extractor(X)
          perceptual_y = vgg19_feature_extractor(Y)
  
          C = perceptual_x.size(0)
          perceptual_x = perceptual_x.reshape(C, -1)
          perceptual_y = perceptual_y.reshape(C, -1)
          cos_dis = np.matmul(perceptual_x.transpose(0, 1), perceptual_y)
          w = np.exp((1 - cos_dis) / band_width)
          cx = w / np.sum(w, axis=2, keepdims=True)
          cx = np.mean(np.max(cx, axis=1)[0], dim=1)
          cx = np.mean(-np.log(cx + 1e-5))
          return cx
  
      # spatial periodic contextual distance
      def SPCX(X, Y, band_width=0.2):
          sub_x = pk_operation(X)
          sub_y = pk_operation(Y)
          
          # ===== different from CX =====
          cos_dis = np.matmul(sub_x, sub_y.transpose(0, 1))
          w = np.exp((1 - cos_dis) / band_width)
          cx = w / np.sum(w, axis=2, keepdims=True)
          cx = np.mean(np.max(cx, axis=1)[0], dim=1)
          cx = np.mean(-np.log(cx + 1e-5))
          return cx
  
      \end{lstlisting}
      \caption{
      Reference numpy code for $\mathcal{P}\mathcal{K}$ operation and spatial periodic contextual distance measurement, as well as the vanilla contextual distance measurement for reference. Here we highlighted the difference between $SPCX$ and $CX$ that denotes context difference. \label{alg:code}}
  \end{table*}
\end{document}